\documentclass{article}

\usepackage{iclr2027_conference,times}

\usepackage{amsmath,amsfonts,bm}

\def\eqref#1{equation~\ref{#1}}

\def\1{\bm{1}}

\DeclareMathAlphabet{\mathsfit}{\encodingdefault}{\sfdefault}{m}{sl}
\SetMathAlphabet{\mathsfit}{bold}{\encodingdefault}{\sfdefault}{bx}{n}

\usepackage{amsmath,amssymb,amsfonts}
\usepackage{array}
\usepackage{booktabs}
\usepackage{algorithm}
\usepackage{algpseudocode}
\usepackage{graphicx}
\usepackage{longtable}
\usepackage{multirow}
\usepackage{ragged2e}
\usepackage{url}
\usepackage{xcolor}
\usepackage{needspace}
\usepackage{placeins}
\usepackage{microtype}
\usepackage{pdflscape}
\usepackage{hyperref}
\hypersetup{hidelinks}
\newcommand{\tblhead}[1]{\textbf{#1}}
\newcolumntype{L}[1]{>{\RaggedRight\arraybackslash}p{#1}}
\newcolumntype{C}[1]{>{\centering\arraybackslash}p{#1}}
\newcommand{\mainTableFormat}{%
  \setlength{\tabcolsep}{4pt}%
  \renewcommand{\arraystretch}{1.08}}
\newcommand{\suppTableFormat}{%
  \setlength{\tabcolsep}{4pt}%
  \renewcommand{\arraystretch}{1.14}}
\newcommand{\suppDenseTableFormat}{%
  \setlength{\tabcolsep}{2.5pt}%
  \renewcommand{\arraystretch}{1.12}}
\newcommand{\tableNoteSep}{\vspace{3pt}}
\newsavebox{\cortivaFigOneBox}
\newcommand{\cortivaFigureOne}{%
  \sbox{\cortivaFigOneBox}{\includegraphics[width=\linewidth]{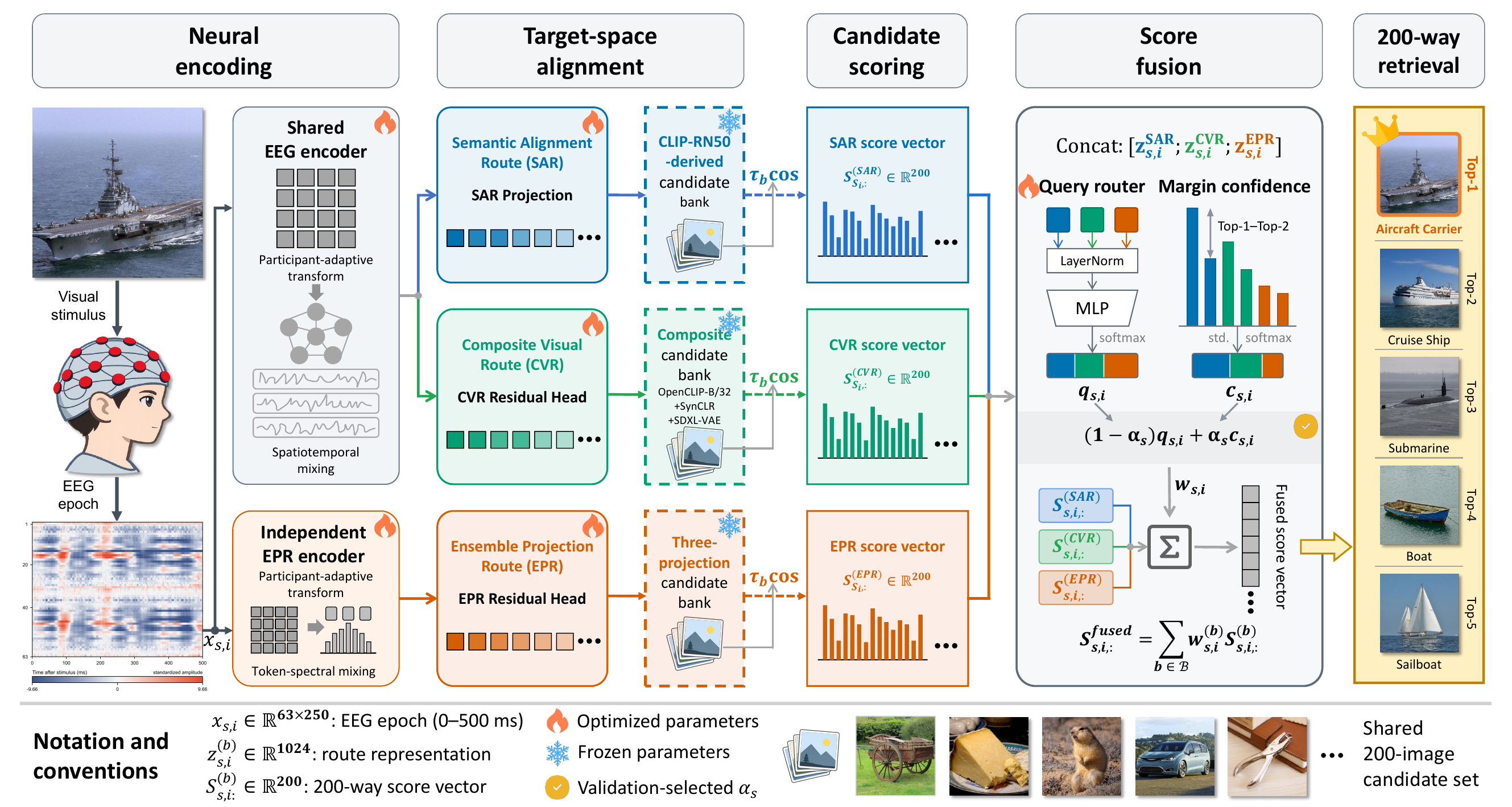}}%
  \usebox{\cortivaFigOneBox}\hspace{-\wd\cortivaFigOneBox}%
  \raisebox{26pt}[0pt][0pt]{\color{white}\rule{\wd\cortivaFigOneBox}{3.0pt}}%
}
\title{CORTIVA: Candidate-Score Fusion of Complementary Visual Teachers for EEG- and MEG-to-Image Retrieval}

\author{Junhan Wang, Kani Chen\thanks{Corresponding author.} \\
The Hong Kong University of Science and Technology \\
Hong Kong SAR, China \\
\texttt{jwangnw@connect.ust.hk} \\
\texttt{makchen@ust.hk}}

\iclrfinalcopy

\begin{document}

\maketitle

\lhead{Under review as a conference paper at ICLR 2027}

\begin{abstract}
Decoding visual experience from non-invasive brain activity is central to neuroscience and brain--computer interfaces. Functional magnetic resonance imaging (fMRI) offers fine spatial detail, but its slow hemodynamics and burdensome acquisition limit temporally resolved decoding. Electroencephalography (EEG) and magnetoencephalography (MEG) provide millisecond resolution, making image retrieval compelling: identify the viewed image from neural responses and a fixed candidate bank. Contrastive alignment to pretrained visual representations enables zero-shot retrieval from EEG and MEG, but most systems collapse heterogeneous visual supervision into a single embedding before ranking. This early consolidation imposes one similarity geometry on every candidate order and removes encoder-specific disagreements from the final ranking. We propose CORTIVA, a candidate-score fusion method that retains this complementary evidence. Three decoding routes are aligned to heterogeneous visual targets, score the same indexed candidates independently, and combine only their temperature-scaled score vectors before ranking. On the 200-way THINGS-EEG2 benchmark, CORTIVA reaches \textbf{73.5\% Top-1 and 95.3\% Top-5} across ten participants, exceeding the strongest reported baseline by \textbf{10.3 and 5.4 percentage points}. With a modality-specific neural encoder, the same fusion principle reaches \textbf{42.4\% Top-1 on THINGS-MEG}. Matched route-removal retraining and four weight controls show that the gain comes from combining complementary route scores and persists under uniform weighting, without depending on a specialized weighting rule. Analyses in an independent DINOv2 space recover the same local error-neighborhood enrichment and posterior neural--visual correspondence. These results support candidate-score fusion as a simple, testable alternative to embedding-level consolidation for neural image retrieval.
\end{abstract}

\section{Introduction}

Neural decoding infers perceptual, cognitive, or motor states from brain activity and supports brain--computer interfaces. Functional magnetic resonance imaging established natural-image decoding by resolving distributed cortical representations \citep{Kay2008NaturalImages,Haxby2001DistributedRepresentations,Shen2019DeepImageReconstruction}, but its slow hemodynamics and burdensome acquisition limit interactive use. EEG and MEG preserve millisecond dynamics \citep{Thorpe1996Speed,Cichy2014ObjectRecognition,Grootswagers2017DecodingTutorial}, motivating participant-level zero-shot image retrieval on public natural-image benchmarks \citep{Gifford2022THINGSEEG2,Song2024NICE,Zhang2026NeuroBridge}.

Recent EEG decoders align neural responses with visual or multimodal embeddings and then rank candidate images in one space \citep{Song2024NICE,Du2023BraVL,Song2025LanguageGuided,Wei2024MB2C,Zhang2025CognitionCapturer,Li2024ATM,Li2025NeuralMCRL,Wu2025UBP}. These systems differ in encoders and supervision, yet a single similarity geometry ultimately orders all candidates. Disagreement among visual targets is therefore compressed before ranking.

CORTIVA instead aligns three routes to heterogeneous visual targets and fuses only their candidate scores. The routes score the same indexed candidate bank independently, so their disagreements remain available when the final ranking is formed. The main contributions are:
\begin{itemize}
\item a retrieval architecture that preserves inter-target ranking differences by combining \textbf{score vectors, not embeddings};
\item \textbf{matched route-removal retraining and weight controls} that separate score integration from weight estimation; and
\item \textbf{participant-level and cross-participant validation across EEG and MEG}, complemented by visual-geometry and neural-similarity analyses.
\end{itemize}

\section{Related Work}

\paragraph{Single-target neural image retrieval.}
THINGS-EEG and THINGS-EEG2 provide public, stimulus-controlled natural-image benchmarks \citep{Grootswagers2022THINGSEEG,Gifford2022THINGSEEG2}. Early decoders classified stimuli within closed label sets using compact networks such as EEGNet \citep{Lawhern2018EEGNet} and deeper spatiotemporal convolutional architectures \citep{Schirrmeister2017DeepLearningEEG}. Contrastive methods reframed the task as retrieval by mapping neural responses into a frozen visual embedding space. NICE aligns EEG with CLIP image embeddings \citep{Song2024NICE,Radford2021CLIP}, and UBP models uncertainty in that alignment \citep{Wu2025UBP}. ATM and Neural-MCRL introduce adaptive and spectral--temporal encoders \citep{Li2024ATM,Li2025NeuralMCRL}, while NeuroBridge combines self-supervised cognitive priors with bidirectional semantic alignment \citep{Zhang2026NeuroBridge}. These systems differ substantially in neural encoding, but one similarity function over one embedding space determines the final candidate ordering.

\paragraph{Structured and multi-source visual targets.}
BraVL combines visual and linguistic features \citep{Du2023BraVL}; NICE++ adds language-guided supervision \citep{Song2025LanguageGuided}; MB2C introduces bidirectional cycle consistency \citep{Wei2024MB2C}; and CognitionCapturer aligns EEG with image, text, and depth modalities \citep{Zhang2025CognitionCapturer}. Candidate targets can also be constructed from supervised backbones, self-supervised representations, or encoders optimized to predict cortical responses \citep{He2016ResNet,Dosovitskiy2021VisionTransformer,Caron2021DINO,He2022MAE,Yamins2014PerformanceOptimized}. Recent methods enrich representations before candidate scoring through structural representations, participant-adaptive granularity, and multilevel alignment \citep{Tang2026HCF,Jiang2026SAMGA,Liu2026MB2L}. In each case, the resulting composite representation still defines one ranking geometry.
We discuss generative reconstruction separately because its perceptual and semantic metrics assess synthesized images rather than rankings within a fixed candidate set \citep{Takagi2023LatentDiffusion,Scotti2024MindEye2,Benchetrit2024BrainDecoding}.

\paragraph{Score-level fusion and multi-teacher distillation.}
Classical classifier-fusion analysis shows that the sum rule can be insensitive to small route-specific posterior errors \citep{Kittler1998ClassifierFusion}, and late score fusion is effective in multimodal semantic recognition \citep{Snoek2005LateFusion}. Gated mixtures of experts instead learn query-dependent weights \citep{Jacobs1991MoE}. CORTIVA applies bidirectional contrastive learning, geometry preservation, and candidate-distribution distillation within each route \citep{Oord2018CPC,Radford2021CLIP,You2017MultiTeacher,Hinton2015Distillation}, but postpones integration until route-specific score vectors are available.

\section{Method}
\label{sec:method}

\paragraph{Datasets.}
THINGS-EEG2 provides 63-channel, 1000-Hz EEG from ten participants, and THINGS-MEG provides 271-sensor, 1200-Hz recordings from four different participants; both use images from the 1,854-concept THINGS collection \citep{Hebart2019THINGS,Gifford2022THINGSEEG2,Hebart2023THINGSData}. EEG trials are baseline-corrected with the \(-100\) to 0 ms interval, cropped to 0--500 ms, and averaged over adjacent sample pairs to 500 Hz, yielding a \(63\times250\) tensor. The primary EEG evaluation averages one test-session mean from each of four sessions. The final 200-image test set is held out from all training, checkpoint selection, and fusion-parameter selection.

\begin{figure}[t]
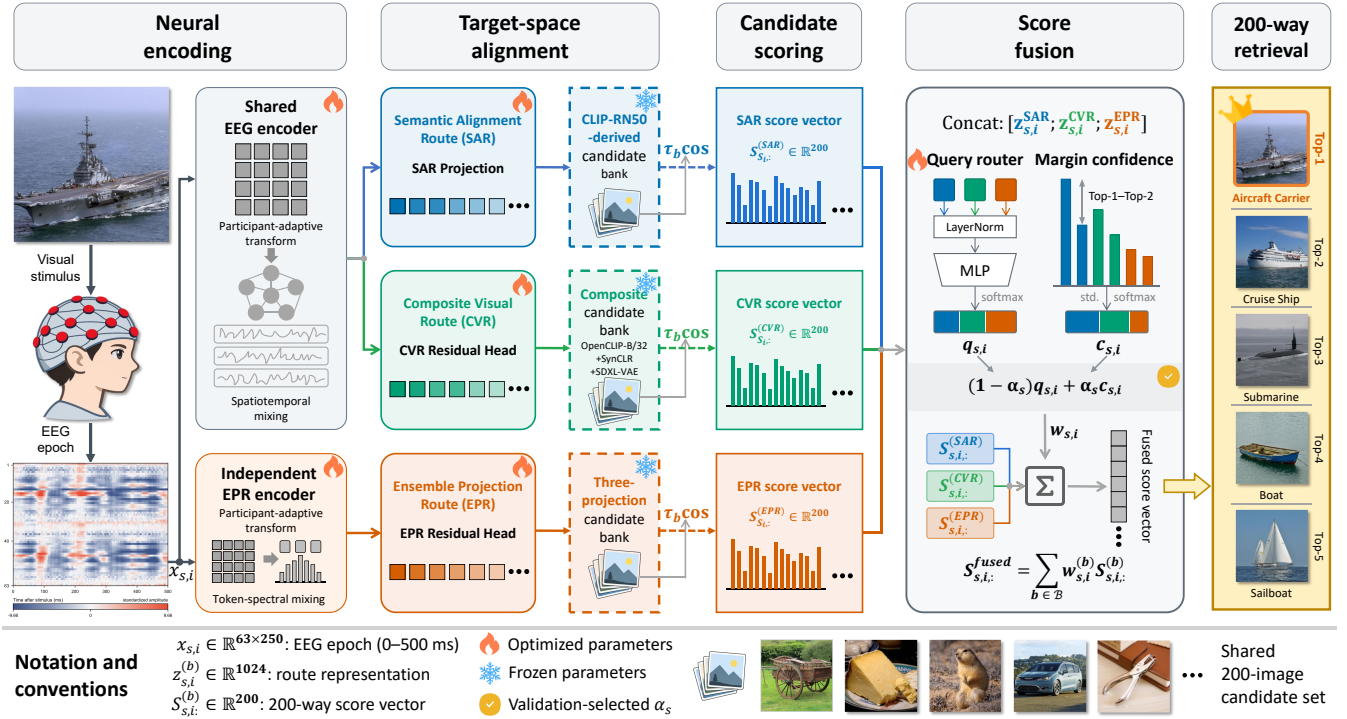

\centering
\cortivaFigureOne
\caption{\textbf{CORTIVA architecture and candidate-score fusion.}
The shared EEG encoder supports SAR and CVR, while EPR uses an independent encoder. Offline visual targets derive from CLIP-RN50, a composite OpenCLIP/SynCLR/SDXL-VAE representation, and a three-projection CLIP-RN50 ensemble. The three identically indexed score vectors are weighted and summed before ranking the common 200-image candidate bank; dashed outlines denote frozen candidate banks.}
\label{fig:cortiva-task-architecture}
\end{figure}

\paragraph{Targets and routes.}
Throughout, a \emph{target} denotes the candidate embedding geometry defined by frozen visual features and fixed transforms; a \emph{route} denotes the neural mapping trained against one target; and a \emph{score vector} contains that route's similarities to the identically indexed candidates. SAR applies a trainable projection to CLIP-RN50 features \citep{Radford2021CLIP}. CVR combines OpenCLIP ViT-B/32, SynCLR, and SDXL-VAE features through fixed source-specific projectors and a fixed composite-target model \citep{Ilharco2021OpenCLIP,Cherti2023OpenCLIP,Tian2024SynCLR,Kingma2014VAE,Podell2023SDXL}. EPR averages three fixed CLIP-RN50 image-projection checkpoints. At inference, the three routes score the same 200-image bank. The complete data flow is summarized in Fig.~\ref{fig:cortiva-task-architecture}.

For participant \(s\), query \(i\), route \(b\in\{\mathrm{sar},\mathrm{cvr},\mathrm{epr}\}\), route representation \(z^{(b)}_{s,i}\), and candidate representation \(v^{(b)}_j\), each route produces
\begin{equation}
S^{(b)}_{sij}=\tau_b
\left\langle
\frac{z^{(b)}_{s,i}}{\lVert z^{(b)}_{s,i}\rVert_2},
\frac{v^{(b)}_j}{\lVert v^{(b)}_j\rVert_2}
\right\rangle ,
\label{eq:route-score}
\end{equation}
where \(\tau_b=\min[\exp(\eta_b),100]\) is a learned inverse temperature. A participant-specific router produces query-dependent route weights and may be blended with a score-margin confidence vector selected on validation data:
\begin{equation}
S^{\mathrm{fused}}_{sij}=\sum_b w^{(b)}_{s,i}S^{(b)}_{sij}.
\label{eq:fused-score}
\end{equation}
Candidates are sorted by descending fused score. The objective combines route-specific bidirectional contrastive losses and distillation terms with cross-entropy on the fused score matrix and router supervision. Loss weights, route definitions, and validation grids were specified before test evaluation; Appendix~\ref{expanded:sec:methods} reports the architecture, objective, and selection procedure.

\paragraph{Neural encoders and selection.}
The shared EEG encoder combines a participant-specific spatial transform, graph attention over a dense directed channel graph, channel-token attention, coordinate-conditioned gating, and a temporal-convolutional projection. SAR uses this projection, CVR applies a residual head, and EPR uses an independent spectral--temporal route. Participant-specific single-target models are trained first; their pairs selected on validation data initialize CORTIVA, after which the target/reference networks are frozen and the route and fusion modules are jointly optimized. Checkpoints are selected by validation loss, and fusion parameters by validation mean rank on concept-disjoint data.

The objective combines route-specific contrastive and distillation terms with cross-entropy for the fused score matrix and router supervision:
\begin{equation}
\mathcal{L}=\mathcal{L}_{\mathrm{SAR}}+\mathcal{L}_{\mathrm{CVR}}+\mathcal{L}_{\mathrm{EPR}}
+0.05\mathcal{L}_{\mathrm{rce}}+0.04\mathcal{L}_{\mathrm{trace}}.
\label{eq:objective}
\end{equation}
The three route losses retain matched-pair contrast, feature direction, within-batch relational geometry, and candidate-score distributions. We optimize encoder-side parameters at $2\times10^{-5}$ and the heads and router at $2\times10^{-4}$ using AdamW with weight decay 0.01, gradient clipping at 1.0, and batches of 384. Training is limited to 60 epochs with patience 5, and the checkpoint with the lowest validation loss is restored before participant-specific fusion parameters are selected by validation mean rank over $\alpha\in\{0,0.1,0.2,0.25,0.3,0.4\}$ and $T_c\in\{2.5,5,10\}$. The held-out test set is scored once after selection.

\paragraph{Computational complexity.}
The trainable Stage 2 set contains 13.8 million parameters per participant. For the Sub01 seed-2026 model on an RTX 4080 SUPER, one query takes 6.30 ms and 92.3 MiB of peak allocated GPU memory when candidate banks are kept on device, excluding offline visual feature extraction. Appendix Table~\ref{tab:supp-computational-footprint} reports the parameter, latency, and memory measurements.

\section{Experimental Protocol}
\label{sec:protocol}

The primary endpoint is participant-level 200-way retrieval on THINGS-EEG2. Each held-out image serves as one query and one candidate; candidates are ranked by descending fused score, with mid-ranks for ties. We report Top-1, Top-5, and mean rank after averaging seeds within participant. Three orthogonal negative controls test candidate identity, post-stimulus timing, and posterior spatial information by permuting candidate labels after scoring, replacing EEG with the duration-matched pre-stimulus interval, or retaining only frontal sensors. Repetition experiments average every combination of one through four test-session means. Selective retrieval orders queries by the fused Top-1-minus-Top-2 margin and reports area under the risk--coverage curve.

For each structural ablation, we retrain the model after removing SAR, CVR, or EPR while retaining the remaining optimization and selection procedure. Four weight controls retain the same 50 participant-by-seed route-score matrices but replace query-aligned weights with within-participant row permutations, participant means, uniform weights, or the fixed prior. Cross-participant experiments initialize each target participant from the best-validation source among the other nine participants and adapt on concept-disjoint target data. The THINGS-MEG configuration uses modality-specific EEGProject and TSConv neural front ends, 512-dimensional route spaces, the full 0--1000 ms epoch, and the same target families and candidate-score fusion rule.

Participant is the unit of inference. Seeds are averaged within participant; percentile-bootstrap intervals summarize endpoints and nonlinear statistics. Paired contrasts use exact sign-flip tests over all $2^{10}$ assignments and 50,000 participant-bootstrap resamples, with Holm adjustment across route-by-metric families. Representational similarity analysis correlates the upper triangles of participant-specific EEG and visual similarity matrices with Spearman $\rho$. DINOv2 ViT-B/14 defines a comparison geometry that is not used by any CORTIVA target. Appendices~\ref{expanded:sec:methods} and~\ref{tab:supp-baseline-protocol} report expanded definitions, hyperparameters, and all comparison sources.

\section{Results}
\label{sec:results}

\begin{figure}[t]
\centering
\includegraphics[width=\linewidth]{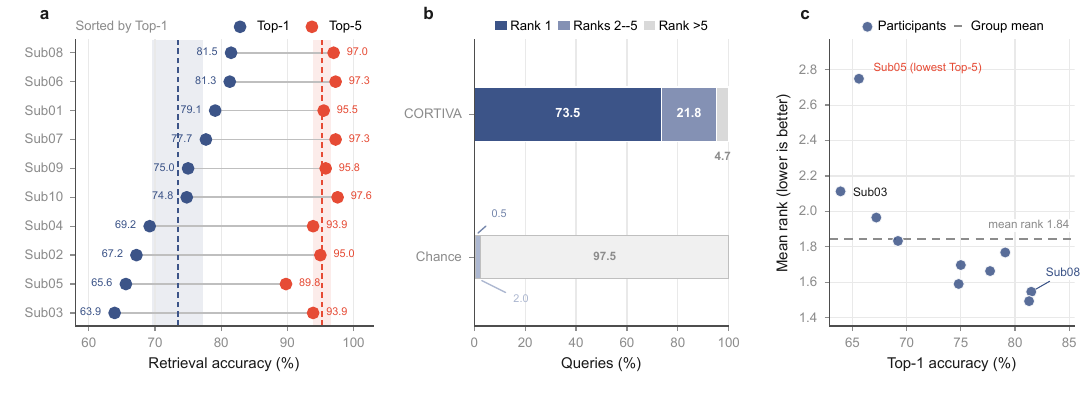}
\caption{\textbf{Participant-level EEG retrieval and rank concentration.}
(a) Five-seed mean Top-1 and Top-5 accuracy for Sub01--Sub10, ordered by Top-1 accuracy; dashed lines mark group means and shaded bands show their 95\% participant-bootstrap intervals. (b) Distribution of the 10,000 participant--seed--query ranks partitioned into rank 1, ranks 2--5, and ranks above 5. (c) Participant-level relation between Top-1 accuracy and mean rank; lower mean rank indicates tighter concentration near the target.}
\label{fig:cortiva-endpoint}
\end{figure}

\paragraph{Within-participant EEG retrieval.}
After averaging seeds within participant, CORTIVA attains \textbf{73.5\% Top-1} accuracy (SD 6.6 percentage points [pp]; 95\% bootstrap CI 69.6--77.3\%) and \textbf{95.3\% Top-5} accuracy (SD 2.4 pp; 95\% CI 93.8--96.6\%) on 200-way THINGS-EEG2 retrieval (Table~\ref{tab:main-results}; Fig.~\ref{fig:cortiva-endpoint}). The strongest reported baseline in the same within-participant setting is NeuroBridge at 63.2\% Top-1 and 89.9\% Top-5 \citep{Zhang2026NeuroBridge}; CORTIVA improves these by \textbf{10.3 and 5.4 pp}.

\begin{table}[t]
\centering
\caption{Participant-level 200-way retrieval on THINGS-EEG2.}
\label{tab:main-results}
\resizebox{\linewidth}{!}{%
\begin{tabular}{@{}ll!{\vrule width 0.4pt}*{10}{r}!{\vrule width 0.4pt}r@{}}
\toprule
\tblhead{Method} & \tblhead{Metric} & \tblhead{Sub01} & \tblhead{Sub02} & \tblhead{Sub03} & \tblhead{Sub04} & \tblhead{Sub05} & \tblhead{Sub06} & \tblhead{Sub07} & \tblhead{Sub08} & \tblhead{Sub09} & \tblhead{Sub10} & \tblhead{Mean} \\
\midrule
\multicolumn{13}{c}{\textbf{Within-participant:} train and test on one participant} \\
\midrule
\multirow{2}{*}{BraVL \citep{Du2023BraVL}} & Top-1 & 6.1 & 4.9 & 5.6 & 5.0 & 4.0 & 6.0 & 6.5 & 8.8 & 4.3 & 7.0 & 5.8 \\
& Top-5 & 17.9 & 14.9 & 17.4 & 15.1 & 13.4 & 18.2 & 20.4 & 23.7 & 14.0 & 19.7 & 17.5 \\
\addlinespace[1.5pt]
\multirow{2}{*}{NICE-GA \citep{Song2024NICE}} & Top-1 & 15.2 & 13.9 & 14.7 & 17.6 & 9.0 & 16.4 & 14.9 & 20.3 & 14.1 & 19.6 & 15.6 \\
& Top-5 & 40.1 & 40.1 & 42.7 & 48.9 & 29.7 & 44.4 & 43.1 & 52.1 & 39.7 & 46.7 & 42.8 \\
\addlinespace[1.5pt]
\multirow{2}{*}{NICE++ w/ GA \citep{Song2025LanguageGuided}} & Top-1 & 16.6 & 20.2 & 19.6 & 25.1 & 13.5 & 16.5 & 17.8 & 26.2 & 18.2 & 22.9 & 19.7 \\
& Top-5 & 46.9 & 47.6 & 52.6 & 61.0 & 42.3 & 48.2 & 51.6 & 62.1 & 48.7 & 53.5 & 51.5 \\
\addlinespace[1.5pt]
\multirow{2}{*}{MB2C \citep{Wei2024MB2C}} & Top-1 & 23.7 & 22.7 & 26.3 & 34.8 & 21.3 & 31.0 & 25.0 & 39.0 & 27.5 & 33.2 & 28.5 \\
& Top-5 & 56.3 & 50.5 & 60.2 & 67.0 & 53.0 & 62.3 & 54.8 & 69.3 & 59.3 & 70.8 & 60.4 \\
\addlinespace[1.5pt]
\multirow{2}{*}{Neural-MCRL \citep{Li2025NeuralMCRL}} & Top-1 & 27.5 & 28.5 & 37.0 & 35.0 & 22.5 & 31.5 & 31.5 & 42.0 & 30.5 & 37.5 & 32.3 \\
& Top-5 & 64.0 & 61.5 & 69.0 & 66.0 & 51.5 & 61.0 & 62.5 & 74.5 & 59.5 & 71.0 & 64.2 \\
\addlinespace[1.5pt]
\multirow{2}{*}{CognitionCapturer \citep{Zhang2025CognitionCapturer}} & Top-1 & 27.2 & 28.7 & 37.2 & 37.7 & 21.8 & 31.6 & 32.8 & 47.6 & 33.4 & 35.1 & 33.3 \\
& Top-5 & 59.5 & 57.0 & 66.1 & 63.2 & 47.8 & 58.1 & 59.6 & 73.5 & 57.6 & 63.6 & 60.6 \\
\addlinespace[1.5pt]
\multirow{2}{*}{UBP \citep{Wu2025UBP}} & Top-1 & 41.2 & 51.2 & 51.2 & 51.1 & 42.2 & 57.5 & 49.0 & 58.6 & 45.1 & 61.5 & 50.9 \\
& Top-5 & 70.5 & 80.9 & 82.0 & 76.9 & 72.8 & 83.5 & 79.9 & 85.8 & 76.2 & 88.2 & 79.7 \\
\addlinespace[1.5pt]
\multirow{2}{*}{NeuroBridge \citep{Zhang2026NeuroBridge}} & Top-1 & 50.0 & 63.2 & 61.6 & 61.4 & 54.8 & 69.7 & 62.7 & 71.2 & 64.0 & 73.6 & 63.2 \\
& Top-5 & 77.6 & 90.6 & 91.1 & 90.0 & 85.0 & 92.9 & 88.8 & 95.1 & 91.0 & 97.1 & 89.9 \\
\addlinespace[1.5pt]
\multirow{2}{*}{\textbf{CORTIVA (Ours)}} & \textbf{Top-1} & \textbf{79.1} & \textbf{67.2} & \textbf{63.9} & \textbf{69.2} & \textbf{65.6} & \textbf{81.3} & \textbf{77.7} & \textbf{81.5} & \textbf{75.0} & \textbf{74.8} & \textbf{73.5} \\
& \textbf{Top-5} & \textbf{95.5} & \textbf{95.0} & \textbf{93.9} & \textbf{93.9} & \textbf{89.8} & \textbf{97.3} & \textbf{97.3} & \textbf{97.0} & \textbf{95.8} & \textbf{97.6} & \textbf{95.3} \\
\midrule
\multicolumn{13}{c}{\textbf{Cross-participant:} leave one participant out for testing} \\
\midrule
\multirow{2}{*}{BraVL \citep{Du2023BraVL}} & Top-1 & 2.3 & 1.5 & 1.4 & 1.7 & 1.5 & 1.8 & 2.1 & 2.2 & 1.6 & 2.3 & 1.8 \\
& Top-5 & 8.0 & 6.3 & 5.9 & 6.7 & 5.6 & 7.2 & 8.1 & 7.6 & 6.4 & 8.5 & 7.0 \\
\addlinespace[1.5pt]
\multirow{2}{*}{NICE \citep{Song2024NICE}} & Top-1 & 7.6 & 5.9 & 6.0 & 6.3 & 4.4 & 5.6 & 5.6 & 6.3 & 5.7 & 8.4 & 6.2 \\
& Top-5 & 22.8 & 20.5 & 22.3 & 20.7 & 18.3 & 22.2 & 19.7 & 22.0 & 17.6 & 28.3 & 21.4 \\
\addlinespace[1.5pt]
\multirow{2}{*}{ATM-S \citep{Li2024ATM}} & Top-1 & 10.5 & 7.1 & 11.9 & 14.7 & 7.0 & 11.1 & 16.1 & 15.0 & 4.9 & 20.5 & 11.84 \\
& Top-5 & 26.8 & 24.8 & 33.8 & 39.4 & 23.9 & 35.8 & 43.5 & 40.3 & 22.7 & 46.5 & 33.73 \\
\addlinespace[1.5pt]
\multirow{2}{*}{UBP \citep{Wu2025UBP}} & Top-1 & 11.5 & 15.5 & 9.8 & 13.0 & 8.8 & 11.7 & 10.2 & 12.2 & 15.5 & 16.0 & 12.4 \\
& Top-5 & 29.7 & 40.0 & 27.0 & 32.3 & 33.8 & 31.0 & 23.8 & 32.2 & 40.5 & 43.5 & 33.4 \\
\addlinespace[1.5pt]
\multirow{2}{*}{Neural-MCRL \citep{Li2025NeuralMCRL}} & Top-1 & 13.0 & 12.0 & 14.5 & 12.5 & 11.5 & 13.5 & 14.0 & 18.5 & 13.5 & 17.0 & 14.0 \\
& Top-5 & 31.5 & 30.5 & 35.5 & 35.5 & 29.0 & 35.5 & 36.0 & 38.5 & 32.5 & 39.0 & 34.3 \\
\addlinespace[1.5pt]
\multirow{2}{*}{NeuroBridge \citep{Zhang2026NeuroBridge}} & Top-1 & 23.2 & 21.2 & 13.2 & 17.0 & 14.5 & 25.0 & 15.3 & 20.1 & 13.7 & 27.2 & 19.0 \\
& Top-5 & 52.4 & 49.3 & 36.5 & 45.3 & 37.7 & 55.0 & 45.1 & 44.9 & 36.5 & 56.3 & 45.9 \\
\addlinespace[1.5pt]
\multirow{2}{*}{\textbf{CORTIVA (Ours)}} & \textbf{Top-1} & \textbf{27.0} & \textbf{26.0} & \textbf{28.2} & \textbf{25.6} & \textbf{27.9} & \textbf{29.0} & \textbf{19.5} & \textbf{32.7} & \textbf{19.2} & \textbf{29.5} & \textbf{26.5} \\
& \textbf{Top-5} & \textbf{57.5} & \textbf{57.3} & \textbf{58.2} & \textbf{64.1} & \textbf{59.5} & \textbf{67.7} & \textbf{51.6} & \textbf{65.5} & \textbf{48.5} & \textbf{64.7} & \textbf{59.5} \\
\bottomrule
\end{tabular}%
}
\vspace{2pt}
\begin{minipage}{0.99\linewidth}
\footnotesize Values are participant means on the 200-way test. CORTIVA reports the five-seed average; analytical chance is 0.5\%/2.5\% for Top-1/Top-5. Supplementary Table~S1 summarizes comparison sources and evaluation settings.
\end{minipage}
\end{table}

Participant Top-1 ranges from 63.9\% (Sub03) to 81.5\% (Sub08), Top-5 from 89.8\% (Sub05) to 97.6\% (Sub10), and mean rank from 1.49 to 2.75 (group mean 1.84). The median rank is 1 for every participant. Performance is stable across seeds at $73.53\pm0.43$\% Top-1, whereas the between-participant SD is 6.6 pp. Validation loss falls by 7.1\% on average when training extends beyond 18 epochs; the matched analyses therefore use the 60-epoch schedule.

\begin{figure}[t]
\centering
\includegraphics[width=\linewidth]{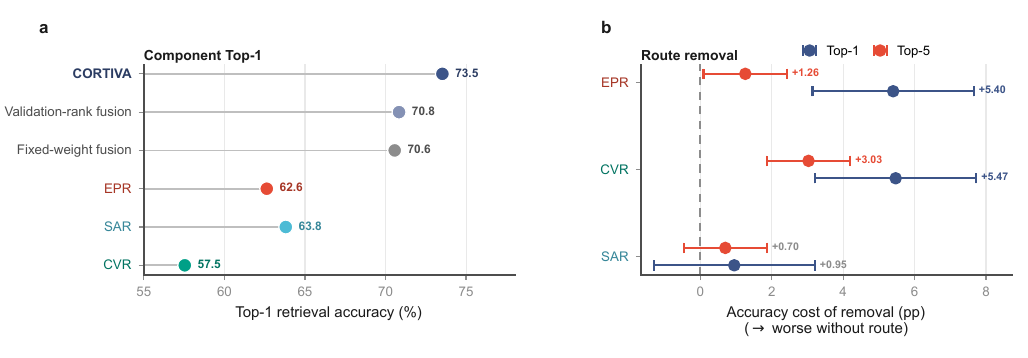}
\caption{\textbf{Matched structural ablations.}
(a) Participant-mean Top-1 retrieval for the three constituent route outputs, two fusion variants, and full CORTIVA. (b) Participant-paired Top-1 and Top-5 accuracy losses after matched retraining without each route; points show full-minus-ablated means, and horizontal bars show simultaneous 95\% participant-bootstrap intervals within each metric. Positive values denote lower accuracy after removal.}
\label{fig:cortiva-ablation}
\end{figure}

\paragraph{Score integration, not a single route.}
CORTIVA exceeds each single-route output in all ten participants: participant means are 9.7 pp above SAR, 16.0 pp above CVR, and 10.9 pp above EPR. Removing CVR and EPR costs 5.47 and 5.40 pp Top-1, respectively, after matched retraining, even though their standalone Top-1 accuracies differ by 5.1 pp (Fig.~\ref{fig:cortiva-ablation}). Across 10,000 participant--seed--query outcomes, fusion ranks the target first in 1,460 cases where EPR does not, while EPR does so in 370 cases where fusion does not. All four route-weight control intervals span zero around the validation-fixed query-aligned rule; uniform summation gives the highest point estimate, 74.2\% versus 73.5\% for the selected query-aligned rule (Appendix Fig.~\ref{expanded:fig:cortiva-router}). These controls indicate that the gain comes from retaining and summing route scores, not from a finely tuned weighting rule.

\paragraph{Cross-participant and MEG transfer.}
Under target-adapted leave-one-participant-out evaluation, CORTIVA reaches \textbf{26.5\% Top-1 and 59.5\% Top-5} on THINGS-EEG2, compared with 19.0\% and 45.9\% for NeuroBridge. With a modality-specific MEG neural encoder and the same target construction and fusion principle, CORTIVA reaches \textbf{42.4\% Top-1 and 73.6\% Top-5} over four THINGS-MEG participants.

MEG participant Top-1 ranges from 20.6\% to 65.6\%, while the seed-wise SD of the four-participant mean is 1.5 pp. In a frozen-checkpoint input analysis, Top-1 rises from 29.2\% with 0--500 ms retained, through 38.5\% with 0--700 ms, to 42.4\% with the full 0--1000 ms epoch. The cumulative profile indicates that useful MEG information is distributed across the full one-second response rather than confined to an early interval.

The complete participant-level comparison is given in Table~\ref{tab:main-meg}.

\begin{table}[H]
\centering
\caption{Within-participant 200-way retrieval on THINGS-MEG.}
\label{tab:main-meg}
\small
\mainTableFormat
\setlength{\tabcolsep}{3pt}
\renewcommand{\arraystretch}{1.02}
\begin{tabular}{@{}ll!{\vrule width 0.4pt}cccc!{\vrule width 0.4pt}c@{}}
\toprule
\tblhead{Method} & \tblhead{Metric} & \tblhead{Sub01} & \tblhead{Sub02} & \tblhead{Sub03} & \tblhead{Sub04} & \tblhead{Mean} \\
\midrule
\multicolumn{7}{c}{\textbf{Within-participant:} train and test on one participant} \\
\midrule
\multirow{2}{*}{NICE \citep{Song2024NICE}} & Top-1 & 9.6 & 18.5 & 14.2 & 9.0 & 12.8 \\
& Top-5 & 27.8 & 47.8 & 41.6 & 26.6 & 36.0 \\
\addlinespace[1.5pt]
\multirow{2}{*}{UBP \citep{Wu2025UBP}} & Top-1 & 15.0 & 46.0 & 27.3 & 18.5 & 26.7 \\
& Top-5 & 38.0 & 80.5 & 59.0 & 43.5 & 55.2 \\
\addlinespace[1.5pt]
\multirow{2}{*}{NeuroBridge \citep{Zhang2026NeuroBridge}} & Top-1 & 16.5 & 53.7 & 40.4 & 18.1 & 32.2 \\
& Top-5 & 41.6 & 85.3 & 73.2 & 43.1 & 60.8 \\
\addlinespace[1.5pt]
\multirow{2}{*}{\textbf{CORTIVA (Ours)}} & \textbf{Top-1} & \textbf{20.6} & \textbf{65.6} & \textbf{52.6} & \textbf{30.9} & \textbf{42.4} \\
& \textbf{Top-5} & \textbf{50.8} & \textbf{91.3} & \textbf{86.5} & \textbf{66.1} & \textbf{73.6} \\
\bottomrule
\end{tabular}
\tableNoteSep
\begin{minipage}{0.98\linewidth}
\footnotesize Sub01--Sub04 denote the MEG cohort. CORTIVA averages seeds 2026--2031; Supplementary Table~S1 summarizes the comparison settings.
\end{minipage}
\end{table}

\paragraph{Ablations, controls, and neural correspondence.}
Matched 60-epoch ablations quantify route-specific effects inside the fused system: removing CVR or EPR produces clear Top-1 losses, whereas the smaller SAR effect is less certain (Fig.~\ref{fig:cortiva-ablation}). Negative controls return retrieval to analytical chance: candidate-label permutation yields 0.50\% Top-1 and 2.50\% Top-5, pre-stimulus input yields 0.60\% and 2.41\%, and frontal-only input yields 0.54\% and 2.41\%. In an independent DINOv2 ViT-B/14 space not used to construct any target, 21.9\% of errors lie among the target's Top-10 neighbors versus 5.0\% at random. EEG--visual representational-similarity analyses show stronger posterior than frontal correspondence, with the posterior-minus-frontal contrast positive in all ten participants.

\begin{figure}[H]
\centering
\includegraphics[width=0.96\linewidth]{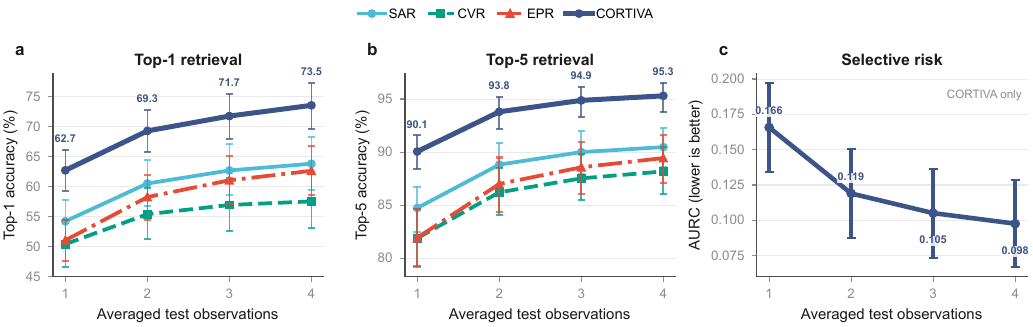}
\caption{\textbf{Retrieval efficiency across available test observations.} \textbf{a,b}, Top-1 and Top-5 retrieval for CORTIVA and its three constituent route outputs after averaging every combination of one to four observations. Markers show participant means, whiskers show 95\% participant-bootstrap intervals, and labels report CORTIVA means. \textbf{c}, CORTIVA AURC under the same observation counts; labels give the four AURC estimates.}
\label{fig:main-repetition}
\end{figure}

\paragraph{Repeated observations and input sensitivity.}
Both retrieval accuracy and confidence ordering improve as additional observations are averaged. From one through four observations, Top-1 rises from 62.7\% through 69.3\% and 71.8\% to 73.5\%, while Top-5 increases from 90.1\% to 95.3\%; over the same range, AURC decreases from 0.166 to 0.098. At the selected checkpoint, retaining only O1, Oz, and O2 yields 29.5\% Top-1. Gaussian noise at 0.05 of the training-set channel SD lowers Top-1 by 1.9 pp. Figure~\ref{fig:main-repetition} shows the complete repetition-efficiency profile; Appendix Sections~\ref{sec:supp-eeg-input} and~\ref{sec:supp-practical-sampling} report the sensor, temporal, perturbation, and repetition details.

\begin{figure}[t]
\centering
\includegraphics[width=0.925\linewidth]{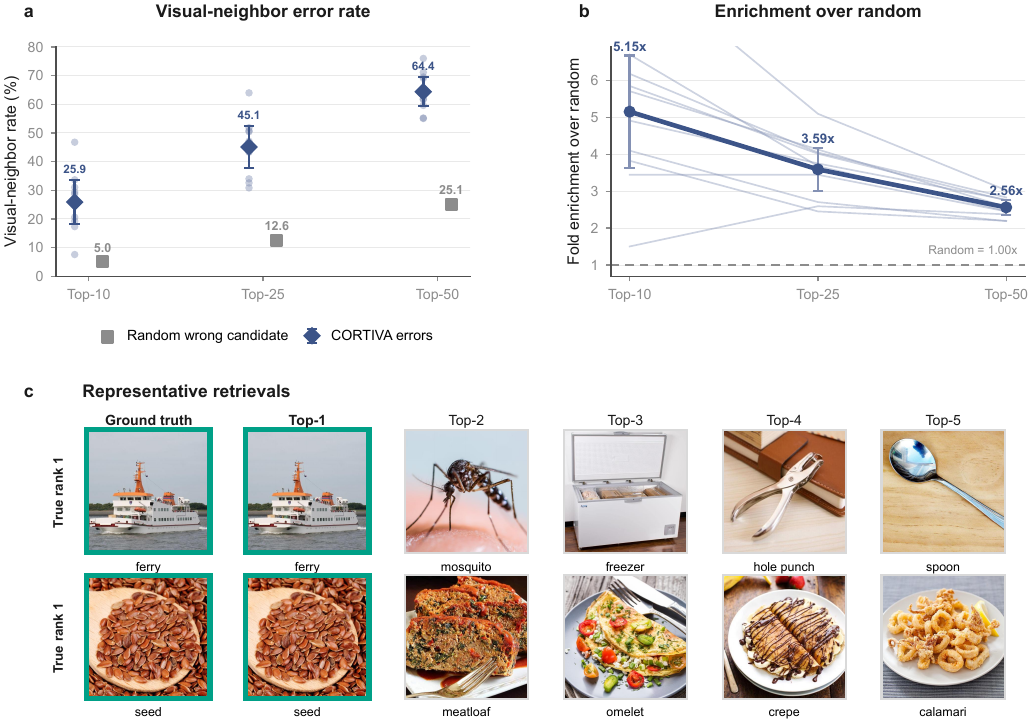}
\caption{\textbf{CLIP-neighbor structure of retrieval errors and real-image examples.}
(a) Participant-level rates at which incorrect CORTIVA predictions fall within the true image's Top-10, Top-25, or Top-50 CLIP neighborhood; squares mark random-wrong-candidate references. (b) Fold enrichment over the corresponding random references, computed within participant before group averaging. (c) Two Top-1 successes from Sub10 (seed 2026), sampled without reference to score or difficulty. Each row shows the ground truth followed by the five highest-ranked candidates; teal borders identify the target.}
\label{fig:cortiva-semantic}
\end{figure}

\begin{figure}[t]
\centering
\includegraphics[width=0.82\linewidth]{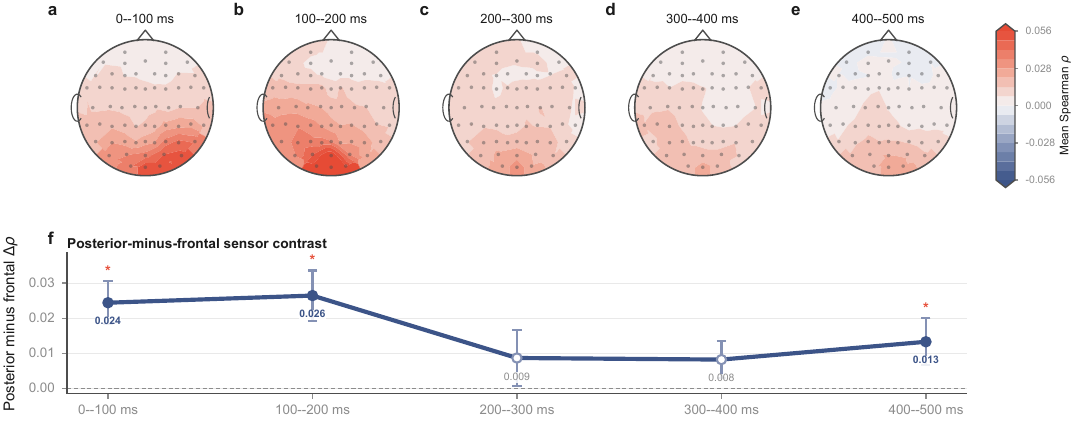}
\caption{\textbf{Spatiotemporal EEG--CLIP representational similarity.}
\textbf{a--e}, Mean single-sensor Spearman correlations for Sub01--Sub10 across five non-overlapping 100-ms windows on one shared asymmetric color scale. Values are interpolated for display only; statistical analyses use the recorded sensor locations.
\textbf{f}, Posterior-minus-frontal contrasts with 95\% \(t\)-intervals across participants; asterisks mark Holm-adjusted one-sided \(p<0.05\) over the five-window family. Regional trajectories appear in Supplementary Fig.~S6.}
\label{fig:main-neural}
\end{figure}

\paragraph{External visual geometry.}
In DINOv2, 21.9\% of incorrect Top-1 predictions lie among the target's Top-10 neighbors versus 5.0\% at random (4.36$\times$; 10/10 participants above random; $p=0.00098$). In SAR's CLIP-RN50 geometry, the corresponding rate is 25.9\% (5.15$\times$ enrichment). The corresponding sensor-wise temporal structure is shown in Fig.~\ref{fig:main-neural}; the error-neighborhood structure is shown in Fig.~\ref{fig:cortiva-semantic}. EEG--DINOv2 RSA shows the same posterior organization as the training-space analysis: mean $\rho=0.0316$ over posterior sensors and a posterior-minus-frontal contrast of $\Delta\rho=0.0108$ (10/10 positive, $p=0.00098$).

\section{Discussion}

\paragraph{Standalone accuracy versus removal cost.}
CVR illustrates why route contribution must be assessed inside the fused system. Its standalone output reaches 57.5\% Top-1, yet removing it costs 5.47 pp, essentially matching the 5.40-pp cost of removing the 62.6\% EPR route. Fusion also recovers 166 outcomes missed at rank one by every constituent route. Full-score complementarity, including useful evidence below rank one, is therefore a more informative basis for route retention than standalone accuracy.

\paragraph{Adaptive weighting and score integration.}
Partly nonoverlapping route successes and near-uniform learned weights (mean normalized entropy 0.938) accord with sum-rule analysis \citep{Kittler1998ClassifierFusion}. Across the four Top-1 controls, intervals span zero and uniform summation gives the highest point estimate. The controls point to score integration as a main source of the gain and suggest uniform summation as a parameter-free alternative.

\paragraph{Modality and participant generalization.}
The score-space formulation extends across modalities and sensor layouts. For EEG, the 6.6-pp between-participant SD is approximately 15 times the 0.43-pp between-seed SD of the group mean. MEG Top-1 spans 45.0 pp, compared with a 17.6-pp EEG range; the seed-wise SD of the MEG group mean is 1.5 pp. Participant-to-participant variation is therefore much larger than optimization-seed variation in both cohorts.

\paragraph{Neural correspondence and limitations.}
Error-neighbor enrichment and RSA measure different properties: the former is local and target-centered, whereas RSA compares all image pairs globally. Local error-neighbor enrichment coexists with modest global EEG--visual correlations, and both patterns also appear in DINOv2, which is not used to construct any target. The posterior-minus-frontal contrast places the stronger association over posterior sensors in both visual spaces. Limitations include candidate-bank retrieval rather than open-ended generation, participant-specific training, and evaluation on public THINGS-family datasets. Broader clinical and online BCI settings remain future work.

\section{Conclusion}

In this article, we introduced CORTIVA, a candidate-score fusion framework that preserves route-specific rankings from three heterogeneous visual teachers until the final decision. On 200-way THINGS-EEG2, CORTIVA reaches \textbf{73.5\% Top-1 and 95.3\% Top-5}, exceeding the strongest reported baseline by 10.3 and 5.4 points; the same principle transfers to THINGS-MEG and cross-participant adaptation. Matched route-removal and weight controls show that full-score complementarity, rather than standalone accuracy or specialized weighting, drives the gain and recovers cases unresolved by every route at rank one. DINOv2 and EEG representational analyses further reveal semantically structured errors and stronger posterior neural--visual correspondence beyond the training targets. CORTIVA therefore establishes candidate-score fusion as an effective and interpretable design principle for non-invasive neural image retrieval.

{\small
\noindent\textbf{Data/code.} THINGS-EEG2 and THINGS-MEG are public datasets \citep{Gifford2021THINGSOSF,Gifford2026THINGSEEG2NeMAR,Hebart2023THINGSData}; code and source data are available at \url{https://github.com/Fuyunhan/CORTIVA}. \textbf{AI use statement.} AI was used only for text translation and polishing; the authors verified the text. \textbf{Ethics.} Source studies report approval and consent \citep{Gifford2022THINGSEEG2,Hebart2023THINGSData}; no identification or clinical deployment is attempted. \textbf{Reproducibility.} The main text gives the protocol; Appendices~\ref{expanded:sec:methods} and~\ref{sec:supp-overview} give details; code is available at \url{https://github.com/Fuyunhan/CORTIVA}.
}

\bibliography{references}
\bibliographystyle{iclr2027_conference}

\clearpage
\appendix

\section{Expanded Background and Positioning}
\label{expanded:sec:background}

\subsection{Introduction}

Neural decoding infers perceptual, cognitive, or motor states from brain activity and supports brain--computer interfaces (BCIs). Early non-invasive systems focused on predefined commands or classes \citep{Lawhern2018EEGNet,Schirrmeister2017DeepLearningEEG}; recent work asks a harder question: which natural image produced a neural response? In retrieval, the decoder ranks a candidate bank to identify that image.

Functional magnetic resonance imaging (fMRI) established visual decoding by resolving distributed cortical representations and enabling natural-image identification and reconstruction \citep{Kay2008NaturalImages,Haxby2001DistributedRepresentations,Shen2019DeepImageReconstruction}. Its seconds-scale hemodynamics and costly, immobile acquisition, however, limit interactive BCIs. Electroencephalography (EEG) and magnetoencephalography (MEG) preserve millisecond dynamics \citep{Thorpe1996Speed,Cichy2014ObjectRecognition,Grootswagers2017DecodingTutorial}, motivating participant-level zero-shot retrieval on public benchmarks \citep{Gifford2022THINGSEEG2,Song2024NICE,Zhang2026NeuroBridge}.

EEG decoding has progressed from closed-set classification to contrastive retrieval of unseen concepts \citep{Song2024NICE,Radford2021CLIP}. Recent systems add language and multimodal targets, uncertainty modeling, and spectral--temporal encoders \citep{Du2023BraVL,Song2025LanguageGuided,Wei2024MB2C,Zhang2025CognitionCapturer,Li2024ATM,Li2025NeuralMCRL,Wu2025UBP}. Yet they ultimately rank candidates in one embedding space. Disagreement among visual sources is therefore compressed before ranking.

CORTIVA instead aligns three routes to heterogeneous visual targets and fuses only their temperature-scaled candidate scores. The routes score the same indexed candidate bank independently, so their disagreements remain available when the final ranking is formed. Matched route-removal retraining and four weight controls test whether the gain comes from score integration rather than a specialized weighting rule. We evaluate the resulting retrieval behavior on THINGS-EEG2 and THINGS-MEG, including participant-level, cross-participant, and visual-geometry analyses \citep{Gifford2022THINGSEEG2,Hebart2019THINGS,Hebart2023THINGSData}.

The main contributions are:
\begin{itemize}
\item a retrieval architecture that scores a shared candidate bank under three heterogeneous visual targets and combines only the resulting score vectors, preserving inter-target ranking differences instead of absorbing them into a single embedding;
\item matched route-removal retraining and four weight controls that quantify each route's removal effect, distinguish score integration from weight estimation, and test full-score complementarity as the basis for route retention; and
\item participant-level and cross-participant validation across EEG and MEG, complemented by a DINOv2 geometry outside target construction that recovers the error-neighborhood and posterior neural-similarity patterns observed in the training spaces.
\end{itemize}

\subsection{Related Work}

\subsubsection{Single-Target Neural Image Retrieval}

THINGS-EEG and THINGS-EEG2 provide public, stimulus-controlled natural-image benchmarks \citep{Grootswagers2022THINGSEEG,Gifford2022THINGSEEG2}. Early decoders classified stimuli within closed label sets using compact networks such as EEGNet \citep{Lawhern2018EEGNet} and deeper spatiotemporal convolutional architectures \citep{Schirrmeister2017DeepLearningEEG}. Contrastive methods reframed the task as retrieval by mapping neural responses into a frozen visual embedding space. NICE aligns EEG with CLIP image embeddings \citep{Song2024NICE,Radford2021CLIP}, and UBP models uncertainty in that alignment \citep{Wu2025UBP}. ATM and Neural-MCRL introduce adaptive and spectral--temporal encoders \citep{Li2024ATM,Li2025NeuralMCRL}, while NeuroBridge combines self-supervised cognitive priors with bidirectional semantic alignment \citep{Zhang2026NeuroBridge}. These systems differ substantially in neural encoding, but one similarity function over one embedding space determines the final candidate ordering.

\subsubsection{Structured and Multi-Source Visual Targets}

BraVL combines visual and linguistic features \citep{Du2023BraVL}; NICE++ adds language-guided supervision \citep{Song2025LanguageGuided}; MB2C introduces bidirectional cycle consistency \citep{Wei2024MB2C}; and CognitionCapturer aligns EEG with image, text, and depth modalities \citep{Zhang2025CognitionCapturer}. More broadly, candidate targets can be constructed from supervised backbones \citep{He2016ResNet,Dosovitskiy2021VisionTransformer}, self-supervised representations \citep{Caron2021DINO,He2022MAE}, or encoders optimized to predict cortical responses \citep{Yamins2014PerformanceOptimized}. In these designs, enrichment occurs upstream of candidate scoring, and the resulting composite representation defines one ranking geometry. Recent methods likewise enrich the representation before candidate scoring through structural representations, participant-adaptive granularity, and multilevel alignment \citep{Tang2026HCF,Jiang2026SAMGA,Liu2026MB2L}. Table~\ref{expanded:tab:cortiva-baselines} summarizes participant-level 200-way retrieval results and their evaluation settings. We discuss generative reconstruction separately because its perceptual and semantic metrics assess synthesized images rather than rankings within a fixed candidate set \citep{Takagi2023LatentDiffusion,Scotti2024MindEye2,Benchetrit2024BrainDecoding}.

\subsubsection{Score-Level Fusion and Multi-Teacher Distillation}

Classical classifier-fusion analysis shows that, when individual posterior estimates deviate only slightly from the true posteriors, the error of the sum rule is insensitive to those deviations to first order \citep{Kittler1998ClassifierFusion}. Late score fusion has also proved effective in multimodal semantic recognition \citep{Snoek2005LateFusion}. A trained combiner adds parameters that must be estimated from finite data, whereas gated mixtures of experts learn query-dependent weights \citep{Jacobs1991MoE}. Each CORTIVA route uses a bidirectional contrastive objective over matched and mismatched image--neural pairs \citep{Oord2018CPC,Radford2021CLIP}; geometry- and candidate-distribution terms connect the objective to multi-teacher distillation \citep{You2017MultiTeacher,Hinton2015Distillation}.

\section{Expanded Methods}
\label{expanded:sec:methods}

\subsection{Dataset and Participants}
\label{expanded:sec:methods-data}

THINGS-EEG2 provides 63-channel, 1000-Hz EEG from ten participants (Sub01--Sub10), while THINGS-MEG provides 271-sensor, 1200-Hz recordings from four different participants; both datasets use images drawn from the 1,854-concept THINGS collection \citep{Hebart2019THINGS,Gifford2022THINGSEEG2,Hebart2023THINGSData}. Models are trained separately for each participant and modality. EEG and MEG participant labels are modality-specific and do not identify the same individuals. Dataset identifiers and acquisition details appear in Supplementary Table~S2.

We start from the released 1000-Hz all-channel arrays. Each trial is baseline-corrected with the \(-100\) to 0 ms interval, cropped to 0--500 ms, and averaged over adjacent sample pairs to 500 Hz, yielding a \(63\times250\) tensor. No additional filtering or artifact rejection is applied before training. For each training image, available session-level trial means are averaged into one input. The primary evaluation averages one trial mean from each of four sessions; a separate repetition analysis scores every combination of one through four test sessions. The 1,654 released concepts are partitioned into 1,454 training and 200 validation concepts (ten images each), producing 14,540 training and 2,000 validation inputs. The held-out set contains one image from each of 200 concepts. Every held-out image serves both as a query and as a candidate, and the entire set is reserved for final evaluation.

\subsection{Visual Targets and Retrieval}
\label{expanded:sec:methods-targets}

Throughout, a \emph{target} denotes the candidate embedding geometry defined by frozen visual features and their fixed transforms; a \emph{route} denotes the neural mapping trained against one target; and a \emph{score vector} contains that route's similarities to the identically indexed candidates. SAR applies a trainable 1,024-d image projection to frozen CLIP-RN50 features \citep{Radford2021CLIP}. For CVR, OpenCLIP ViT-B/32 (512-d), SynCLR (768-d), and SDXL-VAE (1,024-d) features pass through fixed source-specific projectors and a fixed composite-target model, whose normalized output has 1,024 dimensions \citep{Ilharco2021OpenCLIP,Cherti2023OpenCLIP,Schuhmann2022LAION5B,Tian2024SynCLR,Kingma2014VAE,Podell2023SDXL}. EPR applies three fixed image-projection checkpoints to CLIP-RN50 features, normalizes their outputs, and averages them. The CVR and EPR targets and all source feature banks are precomputed; SAR candidate representations are updated through its image projector. At inference, the three routes score fixed candidate banks under the same 200-image index. Supplementary Tables~S3, S8, and~S11 specify the models and checkpoints.

Stage~1 performs participant-specific, single-target EEG--image alignment on the same 1,454 training and 200 validation concepts used by the final model. It produces paired EEG and image-projection checkpoints, including the source-specific CVR projectors and the three EPR projection pairs. Stage~2 initializes CORTIVA from the pairs selected on validation data, freezes the target/reference networks, and jointly optimizes the trainable route and fusion modules described below. The held-out 200-image test set is excluded from both stages.

Let \(b\in\mathcal{B}=\{\mathrm{sar},\mathrm{cvr},\mathrm{epr}\}\) index the three routes, \(z^{(b)}_{s,i}\) denote the route representation for participant \(s\) and held-out query \(i\), and \(v^{(b)}_j\) denote candidate \(j\) in the corresponding route space under the selected model. Each route produces a temperature-scaled cosine score
\begin{equation}
S^{(b)}_{sij}=\tau_b
\left\langle
\frac{z^{(b)}_{s,i}}{\lVert z^{(b)}_{s,i}\rVert_2},
\frac{v^{(b)}_j}{\lVert v^{(b)}_j\rVert_2}
\right\rangle .
\label{expanded:eq:route-score}
\end{equation}
Here \(\tau_b=\min[\exp(\eta_b),100]\) is the learned inverse temperature (logit scale, as in CLIP \citep{Radford2021CLIP}) for route \(b\). The common candidate index makes score-vector positions directly comparable. Fusion preserves each learned score scale; consequently, route influence depends jointly on \(\tau_b\), the score distribution, and the fusion weight rather than on weight magnitude alone.

\subsection{EEG Input and Controls}
\label{expanded:sec:methods-input}

The model input is a \(63\times250\) tensor (channels \(\times\) time samples) spanning 0--500 ms post-stimulus at 500 Hz; the default four-session test average follows Section~\ref{expanded:sec:methods-data}.

Three orthogonal negative controls test candidate indexing, post-stimulus timing, and posterior spatial information: candidate identities are permuted after scoring; the EEG input is replaced by the duration-matched pre-stimulus interval (\(-500\) to 0 ms, all 63 channels); or only frontal sensors are retained over 0--500 ms. Supplementary Fig.~S1 and Table~S13 summarize these controls, and Supplementary Table~S17 reports the sensor and temporal input scan. For selective retrieval, queries are ordered by the fused Top-1-minus-Top-2 score margin; risk is the Top-1 error rate among retained queries, and lower area under the risk--coverage curve (AURC) indicates better confidence ordering. Sensor-space figures display participant-mean statistics at the recorded sensors rather than model weights.

\subsection{CORTIVA Model Family}
\label{expanded:sec:methods-model}

The shared graph-token EEG encoder in Fig.~\ref{fig:cortiva-task-architecture} combines graph-based and token-based channel mixing. A participant-specific \(63\rightarrow63\) spatial transform first remaps the channel axis. The resulting tokens pass through graph attention over a dense directed channel graph without self-loops, followed by channel-token attention. A learned channel gate and an additive bias from a 40-d electrode-coordinate embedding then modulate the output. A temporal-convolutional head projects the representation to 1,024 dimensions. EPR uses a separate participant-specific spatial transform and channel-token Transformer \citep{Vaswani2017Attention}. Its spectral module computes a channel-wise FFT, reweights the coefficients with learned per-channel and per-band gains over five masks (0.5--4, 4--8, 8--13, 13--30, and 30--45 Hz), applies an inverse FFT, and blends the result residually with the input before a separate 1,024-d temporal-convolutional projection. SAR uses the graph-token projection, CVR applies a residual head to that vector, and EPR applies an independent residual head. Supplementary Table~S3 specifies the trainable modules and spectral masks.

Each CVR and EPR residual head contains two 1,024-d linear layers with GELU activation and dropout (\(p=0.1\)) between them. We initialize the final layer at zero, making the residual mapping the identity at initialization. For participant \(s\), router \(g_s(\cdot)\) applies layer normalization to concatenated route vectors followed by a \(3072\!\rightarrow\!512\!\rightarrow\!3\) multilayer perceptron with GELU and 0.05 dropout. Its output weights start at zero, and its bias is initialized to the element-wise logarithm of the prespecified prior \((0.55,0.25,0.20)\), chosen before test evaluation and held fixed across all participants, seeds, and ablations. The router outputs a softmax over the three routes:
\begin{equation}
q_{s,i}=\operatorname{softmax}\!\left(g_s\!\left[
z^{(\mathrm{sar})}_{s,i};
z^{(\mathrm{cvr})}_{s,i};
z^{(\mathrm{epr})}_{s,i}
\right]\right).
\label{expanded:eq:router}
\end{equation}
The router is optionally blended with a score-margin confidence vector \(c_{s,i}\), obtained by standardizing each route's Top-1-minus-Top-2 margin within a query (Supplementary Eq.~(S1)). The participant-specific blend \(\alpha_s\) and confidence temperature \(T_{c,s}\) are chosen on validation data. The final weights and fused score are
\begin{equation}
\begin{aligned}
w_{s,i} &= (1-\alpha_s)q_{s,i}+\alpha_s c_{s,i}, \\
S^{\mathrm{fused}}_{sij} &= \sum_{b\in\mathcal{B}}w^{(b)}_{s,i}S^{(b)}_{sij}.
\end{aligned}
\label{expanded:eq:fused-score}
\end{equation}
Equation~\eqref{expanded:eq:fused-score} implements candidate-score fusion: route scores are combined over aligned candidate indices without collapsing the targets into a common embedding. Candidates are sorted by descending fused score, with ties assigned mid-ranks. For \(N=200\) final queries and candidates, the correct-image rank and participant-level Top-\(k\) accuracy are
\begin{equation}
\begin{aligned}
r_{s,i} &= 1+\sum_{j\neq y_i}\mathbb{1}\!\left[S^{\mathrm{fused}}_{sij}>S^{\mathrm{fused}}_{si y_i}\right] \\
&\quad+\tfrac{1}{2}\sum_{j\neq y_i}\mathbb{1}\!\left[S^{\mathrm{fused}}_{sij}=S^{\mathrm{fused}}_{si y_i}\right], \\
\operatorname{Top\mbox{-}}k_s &= \frac{1}{N}\sum_{i=1}^{N}\mathbb{1}[r_{s,i}\leq k].
\end{aligned}
\label{expanded:eq:topk}
\end{equation}
The objective combines route-specific contrastive and distillation terms with cross-entropy for the fused score matrix and router supervision:
\begin{equation}
\mathcal{L}=\mathcal{L}_{\mathrm{SAR}}+\mathcal{L}_{\mathrm{CVR}}+\mathcal{L}_{\mathrm{EPR}}
+0.05\,\mathcal{L}_{\mathrm{rce}}+0.04\,\mathcal{L}_{\mathrm{trace}}.
\label{expanded:eq:objective}
\end{equation}
Each route loss, \(\mathcal{L}_{\mathrm{SAR}}\), \(\mathcal{L}_{\mathrm{CVR}}\), and \(\mathcal{L}_{\mathrm{EPR}}\), is a weighted sum of a bidirectional contrastive term \citep{Oord2018CPC,Radford2021CLIP} and distillation terms that retain feature direction, within-batch relational geometry, and candidate-score distributions \citep{Hinton2015Distillation,You2017MultiTeacher}. All 15 weights were specified before test evaluation and are shared across participants, seeds, and matched ablations; Supplementary Table~S8 lists the objective and its weights. Contrastive weights are 0.18--0.20 and auxiliary weights are 0.02--0.08.
For a training batch of size \(B\), let \(\widetilde S^{\mathrm{fused}}\in\mathbb{R}^{B\times B}\) denote the within-batch fused score matrix. Routed supervision is symmetric across EEG-to-image and image-to-EEG directions,
\begin{equation}
\mathcal{L}_{\mathrm{rce}}=\tfrac{1}{2}\left[
\operatorname{CE}(\widetilde S^{\mathrm{fused}},I_B)+\operatorname{CE}((\widetilde S^{\mathrm{fused}})^{\mathsf T},I_B)
\right],
\label{expanded:eq:routed-ce}
\end{equation}
where \(I_B=(1,\ldots,B)\) denotes diagonal pairing. Router supervision is derived only from route-specific reference-score margins on training rows. If \(U^{(b)}\) is the reference score matrix for route \(b\), then
\begin{equation}
\begin{aligned}
d_i^{(b)} &= U_{ii}^{(b)}-\max_{j\neq i}U_{ij}^{(b)}, \\
p_i^{(b)} &=
\frac{\exp\!\left(d_i^{(b)}/T_{\mathrm{trace}}\right)}
{\sum_{b'\in\mathcal{B}}\exp\!\left(d_i^{(b')}/T_{\mathrm{trace}}\right)}, \\
\mathcal{L}_{\mathrm{trace}} &= -\frac{1}{B}
\sum_{i=1}^{B}\sum_{b\in\mathcal{B}}p_i^{(b)}\log q_i^{(b)},
\end{aligned}
\label{expanded:eq:router-trace}
\end{equation}
with \(T_{\mathrm{trace}}=5\).

\subsection{Training and Model Selection}
\label{expanded:sec:methods-training}

Stage~2 updates the SAR and EPR projection and feature-normalization blocks, the SAR image projector, both residual heads, the router, and the three logit scales. Earlier backbone blocks, the CVR target model, and all reference networks remain frozen. AdamW \citep{Loshchilov2019AdamW} uses \(\beta=(0.9,0.999)\), weight decay 0.01, gradient clipping at 1.0, batches of 384, and learning rates of \(2\times10^{-5}\) for encoder-side parameters and \(2\times10^{-4}\) for heads and router. Training runs for at most 60 epochs with patience 5, and validation loss selects the checkpoint within that budget.

Model parameters are trained on 1,454 concepts; checkpoints and fusion hyperparameters are selected on 200 disjoint validation concepts. Validation averages the ten images within each concept, whereas testing ranks individual held-out images (Supplementary Table~S9). Because ranks are computed on concept-averaged rows, validation Top-1 is near ceiling and provides little discrimination among checkpoints. Checkpoint and fusion selection therefore use validation loss and mean rank. For \(n_c=10\) validation images per concept, the fused matrix is
\begin{equation}
\overline S^{\mathrm{val}}_{s,cj}(e,\alpha,T_c)
=\frac{1}{n_c}\sum_{h=1}^{n_c}
S^{\mathrm{val}}_{s,(c,h),j}(e,\alpha,T_c),
\label{expanded:eq:validation-aggregation}
\end{equation}
where \((c,h)\) indexes the \(h\)th EEG row for concept \(c\). Equation~\eqref{expanded:eq:validation-aggregation} produces the \(200\times200\) matrix used for checkpoint and fusion selection; participant-specific parameters are listed in Supplementary Table~S9.

For participant \(s\), checkpoint epoch and fusion hyperparameters are selected sequentially from the validation data:
\begin{equation}
\begin{aligned}
e_s^* &= \arg\min_e \mathcal{L}^{\mathrm{val}}_s(e), \\
(\alpha_s^*,T_{c,s}^*) &= \arg\min_{\alpha\in\mathcal{A},\,T_c\in\mathcal{T}}
\frac{1}{200}\sum_{i=1}^{200}r^{\mathrm{val}}_{s,i}(e_s^*,\alpha,T_c),
\end{aligned}.
\label{expanded:eq:validation-selection}
\end{equation}
The blend-weight grid is \(\mathcal{A}=\{0,0.1,0.2,0.25,0.3,0.4\}\), and the confidence-temperature grid is \(\mathcal{T}=\{2.5,5.0,10.0\}\). Mean-rank ties are resolved by Top-1, Top-5, lower \(\alpha\), and lower \(T_c\), in that order. Algorithm~\ref{expanded:alg:cortiva-protocol} summarizes the selection procedure.

\begin{algorithm}[!t]
\caption{CORTIVA training and model selection. Test evaluation follows checkpoint and fusion selection on concept-disjoint validation data.}
\label{expanded:alg:cortiva-protocol}
\begin{algorithmic}[1]
\Require \(\mathcal{D}^{\mathrm{fit}}_s,\mathcal{D}^{\mathrm{val}}_s,\mathcal{D}^{\mathrm{test}}_s\) for \(S\) participants; frozen \(\{V^{(b)}\}_{b\in\mathcal{B}}\); grids \(\mathcal{A},\mathcal{T}\)
\Statex \textbf{Output:} Participant-level retrieval results
\ForAll{participants \(s\in\{1,\ldots,S\}\)}
  \State Initialize the participant-specific encoders, residual heads, and router
  \For{epoch \(e=1,\ldots,60\)}
    \State Update trainable parameters on \(\mathcal{D}^{\mathrm{fit}}_s\) using \eqref{expanded:eq:objective}
    \State Compute \(\mathcal{L}^{\mathrm{val}}_s(e)\); retain the minimum-loss checkpoint
    \State Stop after five epochs without improvement
  \EndFor
  \State Restore the minimum-validation-loss checkpoint \(e_s^*\)
  \ForAll{\((\alpha,T_c)\in\mathcal{A}\times\mathcal{T}\)}
    \State Form \(\overline S^{\mathrm{val}}_s\) by \eqref{expanded:eq:validation-aggregation}; compute ranks by \eqref{expanded:eq:topk}
  \EndFor
  \State Select \((\alpha_s^*,T_{c,s}^*)\) by \eqref{expanded:eq:validation-selection} and the fixed tie rule
  \State Fix \((e_s^*,\alpha_s^*,T_{c,s}^*)\) and score \(\mathcal{D}^{\mathrm{test}}_s\) using \eqref{expanded:eq:route-score}--\eqref{expanded:eq:fused-score}
\EndFor
  \State \Return participant-level Top-1, Top-5, and ranks
\end{algorithmic}
\end{algorithm}

\subsection{Computational Complexity}
\label{expanded:sec:methods-complexity}

The trainable Stage~2 set contains 13.8~M parameters per participant. On an RTX 4080 SUPER, the Sub01 seed-2026 model processes one query in 6.30~ms (median over 1,000 timed iterations) with 92.3~MiB of peak allocated GPU memory when all three candidate banks are on device; offline visual feature extraction is excluded. Supplementary Table~S4 reports parameter, latency, and memory details.

\subsection{Structural Component Ablations}
\label{expanded:sec:methods-ablation}

Three participant-paired variants remove SAR, CVR, or EPR from the fused score. For a removed route, its loss terms are dropped, its fusion weight is set to zero, its router-input slice is zeroed, and its route-exclusive parameters receive no gradient and remain at initialization. All other modules retain the full model's optimization settings. Because CVR depends on the shared EEG encoder, the no-SAR variant retains that encoder and isolates the SAR score vector; the no-CVR and no-EPR variants additionally remove route-exclusive parameters. The three interventions therefore differ in architectural scope while preserving matched optimization and selection.

Each leave-one-route-out model is independently retrained and selected under the same participant-wise evaluation design as full CORTIVA across Sub01--Sub10 and seeds 2026--2030. Participant-paired full-minus-ablated contrasts quantify route-specific effects; the CVR and EPR losses are clear, whereas the smaller SAR effect has a simultaneous interval that spans zero.

Training sufficiency is evaluated by extending otherwise matched runs from 18 to 60 epochs and comparing concept-disjoint validation loss (Supplementary Table~S19).

\subsection{THINGS-MEG Evaluation}
\label{expanded:sec:methods-meg}

THINGS-MEG contains 1,654 training concepts with 12 images each and 200 test images with 12 repetitions each \citep{Hebart2023THINGSData}. For MEG, data are downsampled to 200 Hz, all 12 test repetitions are averaged, route spaces are 512-d, and the selected schedule is refitted on all 1,654 training concepts. The EEG backbones are not reused for MEG; instead, the neural encoder uses the EEGProject and TSConv components introduced by \citet{Zhang2026NeuroBridge}, while retaining CORTIVA's three target constructions and candidate-score fusion rule. Supplementary Table~S5 lists the optimizer, warm-up, and augmentation settings.

For each participant and seed, 1,454 concepts form the training split and 200 form the disjoint selection split. A 15-epoch SAR warm-up precedes joint training for at most 60 epochs with patience 12. Warm-up duration, joint-training duration, and score-margin confidence parameters are chosen on validation data before the model is refitted on all 1,654 concepts using the selected schedule. Results average seeds 2026--2031; Supplementary Tables~S5 and~S6 list the corresponding settings.

\subsection{Cross-Participant Adaptation}
\label{expanded:sec:methods-cross-participant}

For each target participant, the other nine participants provide source models, and the source checkpoint with the lowest validation loss initializes adaptation. Target-participant observations are averaged by concept and divided into concept-disjoint adaptation and validation sets; the held-out retrieval concepts remain disjoint. Final encoder blocks, route heads, router, and target-participant parameters are updated for at most 200 epochs with patience 15; batch-normalization running statistics are then re-estimated on the adaptation split. Participant results average seeds 2026--2030; Supplementary Table~S1 summarizes the evaluation setting.

\subsection{Error and Representational Analyses}
\label{expanded:sec:methods-mechanism}

For each incorrect Top-1 prediction, we record whether the predicted image lies within the correct image's Top-\(k\) neighborhood in SAR's CLIP-RN50 space and compare the resulting rate with the analytical chance rate \(k/199\) for a uniformly drawn wrong candidate. The external analysis repeats this calculation in DINOv2 ViT-B/14, a geometry outside CORTIVA target construction \citep{Oquab2024DINOv2}. Representational similarity analysis (RSA) \citep{Kriegeskorte2008RSA,Cichy2016DNNHierarchy,KhalighRazavi2014DeepModels} correlates the upper triangles of the \(200\times200\) EEG and visual similarity matrices with Spearman \(\rho\), separately for Sub01--Sub10. Posterior sensors have O, PO, P, or CP prefixes; frontal sensors have Fp, AF, or F prefixes; the occipital set is O1/Oz/O2. This analysis uses fixed EEG tensors and repeated measurements, keeping the external-geometry validation separate from model fitting and score selection.

To isolate the value of weight estimation from that of score integration, the controls retain the same 50 participant-by-seed route-score matrices while replacing query-aligned weights with 2,000 within-participant row permutations, participant means, uniform weights, or the \((0.55,0.25,0.20)\) prior. For query \(i\), normalized route-weight entropy is \(-\sum_b w_i^{(b)}\log w_i^{(b)}/\log 3\), ranging from 0 for a single active route to 1 for uniform weights. Max-weight-route accuracy and an any-route Top-1 oracle characterize the range of route-wise performance.

\subsection{Statistics}
\label{expanded:sec:methods-statistics}

Seeds are averaged within participant before group inference. Query-level figures pool seed-level outcomes; inferential tests use participant summaries. Percentile-bootstrap intervals summarize participant variability and nonlinear statistics, while figure captions identify \(t\)-intervals. Paired contrasts use exact sign-flip tests over all \(2^{10}=1{,}024\) assignments \citep{Ernst2004Permutation,Nichols2002PermutationNeuroimaging,Maris2007NonparametricEEGMEG} and 50,000 participant-bootstrap resamples \citep{Efron1993Bootstrap}; endpoint intervals use 100,000 resamples. Posterior-versus-frontal hypotheses use one-sided exact sign tests.

Route-removal effects use simultaneous 95\% participant-bootstrap intervals and Holm adjustment across nine route-by-metric tests \citep{Holm1979MultipleTests}; standardized paired effects are Cohen's \(d_z\). Weight controls use paired percentile-bootstrap intervals with within-metric Holm adjustment. The full-window RSA contrast is primary, five temporal contrasts form a Holm-adjusted family, and 56 time--frequency contrasts use Benjamini--Hochberg correction \citep{Benjamini1995FDR}.

\subsection{Ethics}

This study is a secondary analysis of the public THINGS-EEG2 and THINGS-MEG releases; ethics approval and consent for the original acquisitions are reported by \citet{Gifford2022THINGSEEG2,Hebart2023THINGSData}.

\section{Expanded Results}
\label{expanded:sec:results}

\subsection{Within-Participant Retrieval}
\label{expanded:sec:results-retrieval}

After averaging seeds within participant, CORTIVA attains 73.5\% Top-1 accuracy (SD 6.6 percentage points [pp]; 95\% bootstrap CI 69.6--77.3\%) and 95.3\% Top-5 accuracy (SD 2.4 pp; 95\% CI 93.8--96.6\%) (main-text Fig.~\ref{fig:cortiva-endpoint}; Table~\ref{expanded:tab:cortiva-baselines}). Top-1 ranges from 63.9\% (Sub03) to 81.5\% (Sub08), Top-5 from 89.8\% (Sub05) to 97.6\% (Sub10), and mean rank from 1.49 to 2.75 (group mean 1.84). The median rank is 1 for every participant. Because each participant contributes 200 queries, Top-1 differences are quantized at 0.5 pp.

\clearpage
\begingroup
\scriptsize
\setlength{\tabcolsep}{1.2pt}
\renewcommand{\arraystretch}{0.98}
\begin{longtable}{@{}>{\RaggedRight\arraybackslash}p{0.23\linewidth}l!{\vrule width 0.4pt}*{10}{r}!{\vrule width 0.4pt}r@{}}
\caption{Participant-level 200-way retrieval on THINGS-EEG2: within-participant.}
\label{expanded:tab:cortiva-baselines}\\
\toprule
\tblhead{Method} & \tblhead{Metric} & \tblhead{Sub01} & \tblhead{Sub02} & \tblhead{Sub03} & \tblhead{Sub04} & \tblhead{Sub05} & \tblhead{Sub06} & \tblhead{Sub07} & \tblhead{Sub08} & \tblhead{Sub09} & \tblhead{Sub10} & \tblhead{Mean} \\
\midrule
\endfirsthead
\caption[]{Participant-level 200-way retrieval on THINGS-EEG2 (continued).}\\
\toprule
\tblhead{Method} & \tblhead{Metric} & \tblhead{Sub01} & \tblhead{Sub02} & \tblhead{Sub03} & \tblhead{Sub04} & \tblhead{Sub05} & \tblhead{Sub06} & \tblhead{Sub07} & \tblhead{Sub08} & \tblhead{Sub09} & \tblhead{Sub10} & \tblhead{Mean} \\
\midrule
\endhead
\bottomrule
\endlastfoot
\multicolumn{13}{c}{\textbf{Within-participant:} train and test on one participant} \\
\midrule
\multirow{2}{=}{BraVL \citep{Du2023BraVL}} & Top-1 & 6.1 & 4.9 & 5.6 & 5.0 & 4.0 & 6.0 & 6.5 & 8.8 & 4.3 & 7.0 & 5.8 \\
& Top-5 & 17.9 & 14.9 & 17.4 & 15.1 & 13.4 & 18.2 & 20.4 & 23.7 & 14.0 & 19.7 & 17.5 \\
\addlinespace[1.5pt]
\multirow{2}{=}{NICE-GA \citep{Song2024NICE}} & Top-1 & 15.2 & 13.9 & 14.7 & 17.6 & 9.0 & 16.4 & 14.9 & 20.3 & 14.1 & 19.6 & 15.6 \\
& Top-5 & 40.1 & 40.1 & 42.7 & 48.9 & 29.7 & 44.4 & 43.1 & 52.1 & 39.7 & 46.7 & 42.8 \\
\addlinespace[1.5pt]
\multirow{2}{=}{NICE++ w/ GA \citep{Song2025LanguageGuided}} & Top-1 & 16.6 & 20.2 & 19.6 & 25.1 & 13.5 & 16.5 & 17.8 & 26.2 & 18.2 & 22.9 & 19.7 \\
& Top-5 & 46.9 & 47.6 & 52.6 & 61.0 & 42.3 & 48.2 & 51.6 & 62.1 & 48.7 & 53.5 & 51.5 \\
\addlinespace[1.5pt]
\multirow{2}{=}{MB2C \citep{Wei2024MB2C}} & Top-1 & 23.7 & 22.7 & 26.3 & 34.8 & 21.3 & 31.0 & 25.0 & 39.0 & 27.5 & 33.2 & 28.5 \\
& Top-5 & 56.3 & 50.5 & 60.2 & 67.0 & 53.0 & 62.3 & 54.8 & 69.3 & 59.3 & 70.8 & 60.4 \\
\addlinespace[1.5pt]
\multirow{2}{=}{Neural-MCRL \citep{Li2025NeuralMCRL}} & Top-1 & 27.5 & 28.5 & 37.0 & 35.0 & 22.5 & 31.5 & 31.5 & 42.0 & 30.5 & 37.5 & 32.3 \\
& Top-5 & 64.0 & 61.5 & 69.0 & 66.0 & 51.5 & 61.0 & 62.5 & 74.5 & 59.5 & 71.0 & 64.2 \\
\addlinespace[1.5pt]
\multirow{2}{=}{CognitionCapturer \citep{Zhang2025CognitionCapturer}} & Top-1 & 27.2 & 28.7 & 37.2 & 37.7 & 21.8 & 31.6 & 32.8 & 47.6 & 33.4 & 35.1 & 33.3 \\
& Top-5 & 59.5 & 57.0 & 66.1 & 63.2 & 47.8 & 58.1 & 59.6 & 73.5 & 57.6 & 63.6 & 60.6 \\
\addlinespace[1.5pt]
\multirow{2}{=}{UBP \citep{Wu2025UBP}} & Top-1 & 41.2 & 51.2 & 51.2 & 51.1 & 42.2 & 57.5 & 49.0 & 58.6 & 45.1 & 61.5 & 50.9 \\
& Top-5 & 70.5 & 80.9 & 82.0 & 76.9 & 72.8 & 83.5 & 79.9 & 85.8 & 76.2 & 88.2 & 79.7 \\
\addlinespace[1.5pt]
\multirow{2}{=}{NeuroBridge \citep{Zhang2026NeuroBridge}} & Top-1 & 50.0 & 63.2 & 61.6 & 61.4 & 54.8 & 69.7 & 62.7 & 71.2 & 64.0 & 73.6 & 63.2 \\
& Top-5 & 77.6 & 90.6 & 91.1 & 90.0 & 85.0 & 92.9 & 88.8 & 95.1 & 91.0 & 97.1 & 89.9 \\
\addlinespace[1.5pt]
\multirow{2}{=}{\textbf{CORTIVA (Ours)}} & \textbf{Top-1} & \textbf{79.1} & \textbf{67.2} & \textbf{63.9} & \textbf{69.2} & \textbf{65.6} & \textbf{81.3} & \textbf{77.7} & \textbf{81.5} & \textbf{75.0} & \textbf{74.8} & \textbf{73.5} \\
& \textbf{Top-5} & \textbf{95.5} & \textbf{95.0} & \textbf{93.9} & \textbf{93.9} & \textbf{89.8} & \textbf{97.3} & \textbf{97.3} & \textbf{97.0} & \textbf{95.8} & \textbf{97.6} & \textbf{95.3} \\
\end{longtable}
\endgroup

\begingroup
\scriptsize
\setlength{\tabcolsep}{1.2pt}
\renewcommand{\arraystretch}{0.98}
\begin{longtable}{@{}>{\RaggedRight\arraybackslash}p{0.23\linewidth}l!{\vrule width 0.4pt}*{10}{r}!{\vrule width 0.4pt}r@{}}
\caption{Participant-level 200-way retrieval on THINGS-EEG2: cross-participant.}
\label{expanded:tab:cortiva-baselines-cross}\\
\toprule
\tblhead{Method} & \tblhead{Metric} & \tblhead{Sub01} & \tblhead{Sub02} & \tblhead{Sub03} & \tblhead{Sub04} & \tblhead{Sub05} & \tblhead{Sub06} & \tblhead{Sub07} & \tblhead{Sub08} & \tblhead{Sub09} & \tblhead{Sub10} & \tblhead{Mean} \\
\midrule
\endfirsthead
\caption[]{Participant-level 200-way retrieval on THINGS-EEG2: cross-participant (continued).}\\
\toprule
\tblhead{Method} & \tblhead{Metric} & \tblhead{Sub01} & \tblhead{Sub02} & \tblhead{Sub03} & \tblhead{Sub04} & \tblhead{Sub05} & \tblhead{Sub06} & \tblhead{Sub07} & \tblhead{Sub08} & \tblhead{Sub09} & \tblhead{Sub10} & \tblhead{Mean} \\
\midrule
\endhead
\bottomrule
\endlastfoot
\multirow{2}{=}{BraVL \citep{Du2023BraVL}} & Top-1 & 2.3 & 1.5 & 1.4 & 1.7 & 1.5 & 1.8 & 2.1 & 2.2 & 1.6 & 2.3 & 1.8 \\
& Top-5 & 8.0 & 6.3 & 5.9 & 6.7 & 5.6 & 7.2 & 8.1 & 7.6 & 6.4 & 8.5 & 7.0 \\
\addlinespace[1.5pt]
\multirow{2}{=}{NICE \citep{Song2024NICE}} & Top-1 & 7.6 & 5.9 & 6.0 & 6.3 & 4.4 & 5.6 & 5.6 & 6.3 & 5.7 & 8.4 & 6.2 \\
& Top-5 & 22.8 & 20.5 & 22.3 & 20.7 & 18.3 & 22.2 & 19.7 & 22.0 & 17.6 & 28.3 & 21.4 \\
\addlinespace[1.5pt]
\multirow{2}{=}{ATM-S \citep{Li2024ATM}} & Top-1 & 10.5 & 7.1 & 11.9 & 14.7 & 7.0 & 11.1 & 16.1 & 15.0 & 4.9 & 20.5 & 11.84 \\
& Top-5 & 26.8 & 24.8 & 33.8 & 39.4 & 23.9 & 35.8 & 43.5 & 40.3 & 22.7 & 46.5 & 33.73 \\
\addlinespace[1.5pt]
\multirow{2}{=}{UBP \citep{Wu2025UBP}} & Top-1 & 11.5 & 15.5 & 9.8 & 13.0 & 8.8 & 11.7 & 10.2 & 12.2 & 15.5 & 16.0 & 12.4 \\
& Top-5 & 29.7 & 40.0 & 27.0 & 32.3 & 33.8 & 31.0 & 23.8 & 32.2 & 40.5 & 43.5 & 33.4 \\
\addlinespace[1.5pt]
\multirow{2}{=}{Neural-MCRL \citep{Li2025NeuralMCRL}} & Top-1 & 13.0 & 12.0 & 14.5 & 12.5 & 11.5 & 13.5 & 14.0 & 18.5 & 13.5 & 17.0 & 14.0 \\
& Top-5 & 31.5 & 30.5 & 35.5 & 35.5 & 29.0 & 35.5 & 36.0 & 38.5 & 32.5 & 39.0 & 34.3 \\
\addlinespace[1.5pt]
\multirow{2}{=}{NeuroBridge \citep{Zhang2026NeuroBridge}} & Top-1 & 23.2 & 21.2 & 13.2 & 17.0 & 14.5 & 25.0 & 15.3 & 20.1 & 13.7 & 27.2 & 19.0 \\
& Top-5 & 52.4 & 49.3 & 36.5 & 45.3 & 37.7 & 55.0 & 45.1 & 44.9 & 36.5 & 56.3 & 45.9 \\
\addlinespace[1.5pt]
\multirow{2}{=}{\textbf{CORTIVA (Ours)}} & \textbf{Top-1} & \textbf{27.0} & \textbf{26.0} & \textbf{28.2} & \textbf{25.6} & \textbf{27.9} & \textbf{29.0} & \textbf{19.5} & \textbf{32.7} & \textbf{19.2} & \textbf{29.5} & \textbf{26.5} \\
& \textbf{Top-5} & \textbf{57.5} & \textbf{57.3} & \textbf{58.2} & \textbf{64.1} & \textbf{59.5} & \textbf{67.7} & \textbf{51.6} & \textbf{65.5} & \textbf{48.5} & \textbf{64.7} & \textbf{59.5} \\
\end{longtable}
\tableNoteSep
\begin{minipage}{0.99\linewidth}
Values are participant means on the 200-way test. CORTIVA reports the five-seed average; analytical chance is 0.5\%/2.5\% for Top-1/Top-5. Supplementary Table~S1 summarizes comparison sources and evaluation settings.
\end{minipage}
\endgroup
\clearpage

\subsection{Stability Across Seeds and Training Duration}
\label{expanded:sec:results-transfer}

Participant-mean Top-1 is stable across seeds at \(73.53\pm0.43\)\%, whereas between-participant variability is larger (SD 6.6 pp; Supplementary Table~S20). Extending training beyond 18 epochs reduces participant-wise validation loss by 7.1\% on average, supporting the 60-epoch schedule used throughout the matched analysis (Supplementary Table~S19).

\subsection{Structural Component Ablations}
\label{expanded:sec:results-ablation}

The frozen EPR projection reference reaches 65.0\% Top-1 before final CORTIVA training (Supplementary Table~S14), and full CORTIVA exceeds it by 8.5 pp. Removing CVR and EPR produces Top-1 losses of 5.47 and 5.40 pp, respectively, each positive in all ten participants (main-text Fig.~\ref{fig:cortiva-ablation}; Supplementary Table~S15). Their single-route Top-1 accuracies differ by 5.1 pp (57.5\% for CVR and 62.6\% for EPR), so standalone accuracy and removal cost do not order the two routes in the same way.

\begin{figure*}[!t]
\centering
\includegraphics[width=\linewidth]{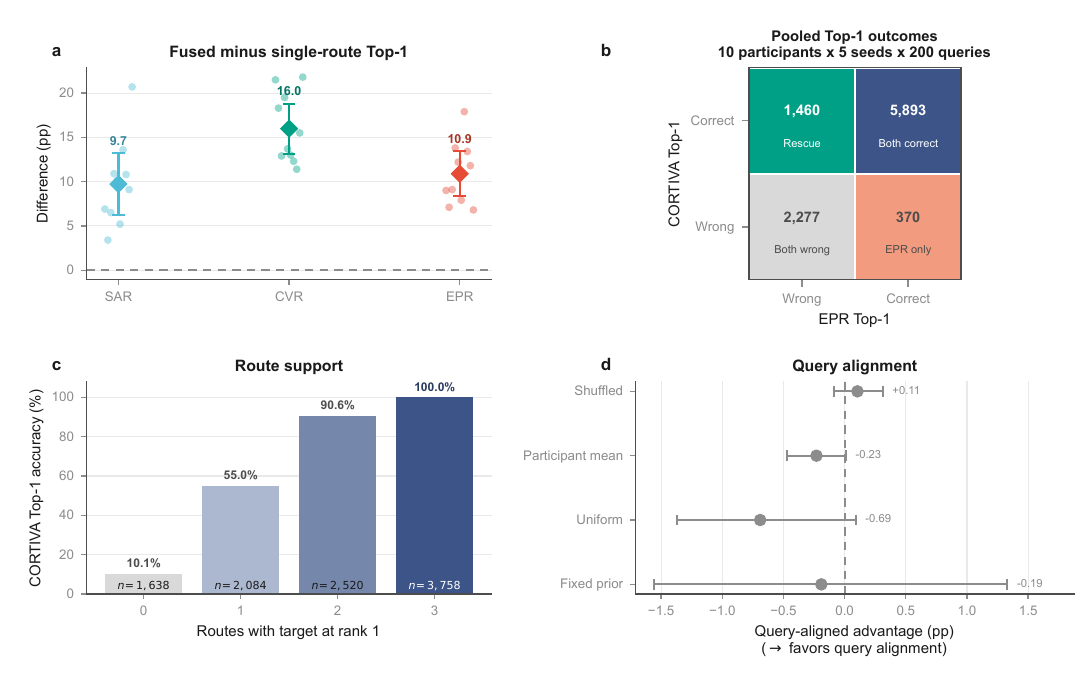}
\caption{\textbf{Score integration and route-weight controls.}
\textbf{a}, Participant-paired fused-minus-single-route Top-1 differences; diamonds and whiskers show means and 95\% \(t\)-intervals.
\textbf{b}, Top-1 outcome transitions between CORTIVA and EPR across all evaluated participant--seed--query combinations.
\textbf{c}, CORTIVA Top-1 accuracy conditioned on how many routes individually rank the correct image first; bar labels give outcome counts. The zero-route column contains images recovered at rank 1 by score-weighted summation from sub-Top-1 route evidence.
\textbf{d}, Participant-paired Top-1 advantages of query-aligned weights over four controls after averaging five seeds within participant. Whiskers are 95\% participant-bootstrap intervals; positive values favor query alignment.}
\label{expanded:fig:cortiva-router}
\end{figure*}

CORTIVA exceeds each single-route output in all ten participants (Fig.~\ref{expanded:fig:cortiva-router}a): participant means are 9.7 pp above SAR, 16.0 pp above CVR, and 10.9 pp above EPR. Together with the matched route-removal effects, these gains show that score integration combines route-specific evidence rather than inheriting the strongest route. Across 10,000 participant~\(\times\)~seed~\(\times\)~query outcomes, CORTIVA ranks the target first in 1,460 cases where EPR does not, while EPR ranks it first in 370 cases where CORTIVA does not (Fig.~\ref{expanded:fig:cortiva-router}b). Conditioning on the number of individually correct routes, CORTIVA's Top-1 accuracy rises from 10.1\% (zero routes) to 55.0\% (one), 90.6\% (two), and 100.0\% (three). Thus, agreement among all routes is preserved, while fusion recovers 166 of the 1,638 outcomes in which no route ranks the target first.

All four Top-1 control intervals span zero around the validation-fixed query-aligned rule (Supplementary Table~S16). Uniform summation gives the highest point estimate, 74.2\% versus 73.5\% for query-aligned weighting (query-aligned minus uniform: \(-0.69\) pp, 95\% CI \([-1.37,0.09]\); Fig.~\ref{expanded:fig:cortiva-router}d). The validation-fixed rule defines the primary result, and uniform summation provides a parameter-free deployment configuration.

Concept-averaged validation Top-1 ranges from 98.5\% to 100.0\%, and eight of ten participants select \(\alpha_s=0\) for every seed (Supplementary Table~S9). Mean rank supplies the finer selection signal under this ceiling-level Top-1 accuracy.

\subsection{Cross-Participant Adaptation}
\label{expanded:sec:results-cross-participant}

Under target-adapted leave-one-participant-out evaluation, CORTIVA achieves 26.5\% Top-1 and 59.5\% Top-5 across five seeds. Participant means range from 19.2\% to 32.7\% and from 48.5\% to 67.7\%, respectively. Table~\ref{expanded:tab:cortiva-baselines-cross} presents the cross-participant results; Supplementary Table~S1 summarizes the comparison settings.

\subsection{Internal Fusion Variants}
\label{expanded:sec:results-baselines}

Fixed-weight and validation-rank fusion reach 70.6\% and 70.8\% Top-1, respectively, compared with 73.5\% for CORTIVA (Supplementary Table~S14).

\FloatBarrier
\subsection{Retrieval on THINGS-MEG}
\label{expanded:sec:results-meg}

\begin{table}[!t]
\centering
\caption{Within-participant 200-way retrieval on THINGS-MEG.}
\label{expanded:tab:cortiva-things-meg}
\mainTableFormat
\setlength{\tabcolsep}{4.5pt}
\begin{tabular}{@{}ll!{\vrule width 0.4pt}cccc!{\vrule width 0.4pt}c@{}}
\toprule
\tblhead{Method} & \tblhead{Metric} & \tblhead{Sub01} & \tblhead{Sub02} & \tblhead{Sub03} & \tblhead{Sub04} & \tblhead{Mean} \\
\midrule
\multicolumn{7}{c}{\textbf{Within-participant:} train and test on one participant} \\
\midrule
\multirow{2}{*}{NICE \citep{Song2024NICE}} & Top-1 & 9.6 & 18.5 & 14.2 & 9.0 & 12.8 \\
& Top-5 & 27.8 & 47.8 & 41.6 & 26.6 & 36.0 \\
\addlinespace[1.5pt]
\multirow{2}{*}{UBP \citep{Wu2025UBP}} & Top-1 & 15.0 & 46.0 & 27.3 & 18.5 & 26.7 \\
& Top-5 & 38.0 & 80.5 & 59.0 & 43.5 & 55.2 \\
\addlinespace[1.5pt]
\multirow{2}{*}{NeuroBridge \citep{Zhang2026NeuroBridge}} & Top-1 & 16.5 & 53.7 & 40.4 & 18.1 & 32.2 \\
& Top-5 & 41.6 & 85.3 & 73.2 & 43.1 & 60.8 \\
\addlinespace[1.5pt]
\multirow{2}{*}{\textbf{CORTIVA (Ours)}} & \textbf{Top-1} & \textbf{20.6} & \textbf{65.6} & \textbf{52.6} & \textbf{30.9} & \textbf{42.4} \\
& \textbf{Top-5} & \textbf{50.8} & \textbf{91.3} & \textbf{86.5} & \textbf{66.1} & \textbf{73.6} \\
\bottomrule
\end{tabular}
\tableNoteSep
\begin{minipage}{0.98\linewidth}
Sub01--Sub04 denote the MEG cohort. CORTIVA averages seeds 2026--2031; Supplementary Table~S1 summarizes the comparison settings.
\end{minipage}
\end{table}

The MEG instantiation reuses the EEGProject and TSConv neural-front-end components introduced by NeuroBridge while retaining CORTIVA's target construction and score fusion. Across six seeds, CORTIVA attains 42.4\% Top-1 and 73.6\% Top-5. Participant Top-1 ranges from 20.6\% (Sub01) to 65.6\% (Sub02), while the seed-wise SD of the four-participant mean is 1.5 pp (Supplementary Table~S6). In the frozen-checkpoint input analysis, Top-1 rises monotonically as the retained interval expands: 29.2\% for 0--500 ms, 38.5\% for 0--700 ms, and 42.4\% for 0--1000 ms. The cumulative profile indicates useful information across the full 1-s epoch (Supplementary Fig.~S8).

\Needspace{4\baselineskip}
\subsection{Repeated-Observation Efficiency and Input Sensitivity}
\label{expanded:sec:results-sensitivity}

Both retrieval accuracy and confidence ordering improve as additional observations are averaged: from one to four repetitions, Top-1 rises from 62.7\% through 69.3\% and 71.8\% to 73.5\%, while Top-5 increases from 90.1\% to 95.3\% (main-text Fig.~\ref{fig:main-repetition}). Over the same range, AURC decreases from 0.166 to 0.098.

All three negative controls return retrieval to analytical chance (Supplementary Fig.~S1; Table~S13): candidate-label permutation yields 0.50\% Top-1 and 2.50\% Top-5, pre-stimulus input yields 0.60\% and 2.41\%, and frontal-only input yields 0.54\% and 2.41\%, respectively. At the selected checkpoint, retaining only O1, Oz, and O2 yields 29.5\% Top-1. Additive Gaussian noise at 0.05 of the training-set channel SD lowers Top-1 from 73.5\% to 71.6\% (\(-1.9\) pp); Supplementary Figs.~S3 and~S4 show the complete sensor, temporal, and perturbation profiles.

\subsection{Visual Geometry of Errors and EEG Responses}
\label{expanded:sec:results-mechanism}

The local error structure extends to a visual space not used to construct any target. In DINOv2 ViT-B/14, 21.9\% of errors lie among the target's Top-10 neighbors versus 5.0\% at random (4.36\(\times\); 10/10 participants above random; one-sided exact sign test, \(p=0.00098\); Supplementary Table~S21). In SAR's CLIP-RN50 training geometry, incorrect Top-1 predictions fall within the correct image's Top-10 neighborhood in 25.9\% of cases, compared with 5.0\% for a random wrong candidate (5.15\(\times\) enrichment). The corresponding Top-25 and Top-50 rates are 45.1\% versus 12.6\% and 64.4\% versus 25.1\% (2.56\(\times\) at Top-50; main-text Fig.~\ref{fig:cortiva-semantic}).

Correlations between participant-level EEG and CLIP similarity matrices are consistently positive: the mean Spearman \(\rho\) over posterior sensors is 0.047, positive in all ten participants, and the occipital group (O1, Oz, and O2) has a full-window mean of \(\rho=0.064\). The full-window posterior-minus-frontal contrast is \(\Delta\rho=0.027\) (10/10 positive; one-sided exact sign test, \(p=0.00098\)). Among the five 100-ms windows, the largest descriptive contrast occurs at 100--200 ms (\(\Delta\rho=0.026\)); contrasts remain positive after Holm correction at 0--100, 100--200, and 400--500 ms. EEG--DINOv2 RSA shows the same organization, with mean \(\rho=0.0316\) over posterior sensors and a posterior-minus-frontal contrast of \(\Delta\rho=0.0108\) (10/10 positive; \(p=0.00098\); Supplementary Table~S21). Supplementary Figs.~S6 and~S7 detail the CLIP-based temporal and spectral structure.

\section{Expanded Discussion}
\label{expanded:sec:discussion}

\subsection{Standalone Accuracy versus Removal Cost}

CVR illustrates why route contribution must be assessed inside the fused system. Its standalone output reaches 57.5\% Top-1, yet removing it costs 5.47 pp, essentially matching the 5.40-pp cost of removing the 62.6\% EPR route. Fusion also recovers 166 outcomes missed at rank one by every constituent route. Full-score complementarity, including useful evidence below rank one, is therefore a more informative basis for route retention than standalone accuracy.

\subsection{Adaptive Weighting and Score Integration}

Partly nonoverlapping route successes and near-uniform learned weights (mean normalized entropy 0.938) accord with the sum-rule analysis of \citet{Kittler1998ClassifierFusion}. Across the four Top-1 controls, intervals span zero and uniform summation gives the highest point estimate. The controls point to score integration as a main source of the gain and suggest uniform summation as a parameter-free alternative.

\subsection{Modality and Participant Generalization}

The score-space formulation extends across modalities and sensor layouts: it reaches 42.4\% Top-1 with 271-sensor MEG and 26.5\% under target-adapted cross-participant evaluation. For EEG, the 6.6-pp between-participant SD is approximately 15 times the 0.43-pp between-seed SD of the group mean. MEG Top-1 spans 45.0 pp (20.6--65.6\%), compared with a 17.6-pp EEG range (63.9--81.5\%); the seed-wise SD of the MEG group mean is 1.5 pp. Participant-to-participant variation is therefore much larger than optimization-seed variation in both cohorts.

\Needspace{4\baselineskip}
\subsection{Neural Correspondence}

Error-neighbor enrichment and RSA measure different properties. The former is local and target-centered, whereas RSA compares all image pairs globally. Local error-neighbor enrichment coexists with modest global EEG--visual correlations, and both patterns also appear in DINOv2, which is not used to construct any target. The posterior-minus-frontal contrast places the stronger association over posterior sensors in both visual spaces.

\subsection{Design Implications}

The results favor retaining target-specific candidate scores through the final stage of neural retrieval. This design adds one candidate-score vector per target, supports direct route-removal tests, and does not require query-dependent weighting. Evaluation in DINOv2 separates the observed error organization from the visual spaces used to train the routes.

\section{Expanded Conclusion}

CORTIVA retains route-specific rankings from three heterogeneous visual teachers until final candidate scoring. On 200-way THINGS-EEG2, it reaches 73.5\% Top-1 and 95.3\% Top-5, exceeding the strongest reported baseline by 10.3 and 5.4 points; the same principle transfers to THINGS-MEG and cross-participant adaptation. Matched route-removal and weight controls show that full-score complementarity, not standalone accuracy or specialized weighting, drives the gain and recovers cases unresolved by every route at rank one. DINOv2 and EEG representational analyses show semantically structured errors and stronger posterior neural--visual correspondence beyond the training targets. Together, these findings support candidate-score fusion as an effective and interpretable design for non-invasive neural image retrieval.

\Needspace{4\baselineskip}

\clearpage
\setcounter{section}{0}
\setcounter{table}{0}
\setcounter{figure}{0}
\setcounter{equation}{0}
\renewcommand{\thesection}{S\arabic{section}}
\renewcommand{\thetable}{S\arabic{table}}
\renewcommand{\thefigure}{S\arabic{figure}}
\renewcommand{\theequation}{S\arabic{equation}}
\renewcommand{\theHsection}{supp.\arabic{section}}
\renewcommand{\theHtable}{supp.\arabic{table}}
\renewcommand{\theHfigure}{supp.\arabic{figure}}
\renewcommand{\theHequation}{supp.\arabic{equation}}
\section{Supplementary overview}
\label{sec:supp-overview}

All group-level retrieval, ablation, routing, and representational-similarity analyses use the participant as the unit of inference. Query-level counts summarize 10,000 participant~\(\times\)~seed~\(\times\)~query outcomes, while inferential tests use participant-level summaries. Percentile-bootstrap intervals summarize endpoints, paired effects, and nonlinear statistics; figures explicitly marked as descriptive means use \(t\)-intervals. THINGS-MEG results aggregate four participants over six optimization seeds. Figures use a color-vision-deficiency-safe palette with redundant marker and line-style encodings where applicable.

\section{Comparison sources and settings}

Table~\ref{tab:supp-baseline-protocol} summarizes the source and evaluation setting of each comparison. Participant entries and aggregate means follow the cited reports. In the NICE family, GA denotes graph attention; CognitionCapturer refers to its visual-output setting.

\begingroup
\suppTableFormat
\renewcommand{\arraystretch}{1.08}
\begin{longtable}{@{}L{0.24\textwidth}L{0.18\textwidth}L{0.31\textwidth}L{0.19\textwidth}@{}}
\caption{Sources and evaluation settings for 200-way results.}\label{tab:supp-baseline-protocol}\\
\toprule
\tblhead{Method} & \tblhead{Publication} & \tblhead{Result source} & \tblhead{Evaluation setting} \\
\midrule
\endfirsthead
\caption[]{Sources and evaluation settings for 200-way results (Continued).}\\
\toprule
\tblhead{Method} & \tblhead{Publication} & \tblhead{Result source} & \tblhead{Evaluation setting} \\
\midrule
\endhead
\bottomrule
\endlastfoot
\multicolumn{4}{l}{\textbf{A. THINGS-EEG2 within-participant retrieval}} \\
\midrule
BraVL \citep{Du2023BraVL} & IEEE TPAMI 2023 & Table VIII & Within participant \\
NICE-GA \citep{Song2024NICE} & ICLR 2024 & Table 2 & Within participant \\
NICE++ w/ GA \citep{Song2025LanguageGuided} & IEEE TNNLS 2025 & Table III & Within participant \\
MB2C \citep{Wei2024MB2C} & ACM MM 2024 & Table 1 & Within participant \\
Neural-MCRL \citep{Li2025NeuralMCRL} & IEEE ICME 2025 & Table I & Within participant \\
CognitionCapturer \citep{Zhang2025CognitionCapturer} & AAAI 2025 & Table 2, image branch & Within participant \\
UBP \citep{Wu2025UBP} & CVPR 2025 & Table 1 & Within participant \\
NeuroBridge \citep{Zhang2026NeuroBridge} & AAAI 2026 & Table 1 & Within participant \\
\textbf{CORTIVA} & Current study & Main Table 1 & Within participant \\
\midrule
\multicolumn{4}{l}{\textbf{B. THINGS-EEG2 cross-participant retrieval}} \\
\midrule
BraVL \citep{Du2023BraVL} & IEEE TPAMI 2023 & Table VIII, 200-way V\&T & Cross participant \\
NICE \citep{Song2024NICE} & ICLR 2024 & Table 2 & Cross participant \\
ATM-S \citep{Li2024ATM}$^{\dagger}$ & NeurIPS 2024 & UBP Table 1; NeuroBridge Table 1 & Cross participant \\
UBP \citep{Wu2025UBP} & CVPR 2025 & Table 1 & Cross participant \\
Neural-MCRL \citep{Li2025NeuralMCRL} & IEEE ICME 2025 & Table I & Cross participant \\
NeuroBridge \citep{Zhang2026NeuroBridge} & AAAI 2026 & Table 1 & Cross participant \\
\textbf{CORTIVA} & Current study & Main Table 1 & Cross participant \\
\midrule
\multicolumn{4}{l}{\textbf{C. THINGS-MEG within-participant retrieval}} \\
\midrule
NICE \citep{Song2024NICE} & ICLR 2024 & Appendix Table 8, retrieval & Within participant \\
UBP \citep{Wu2025UBP} & CVPR 2025 & Table 2, intra-subject & Within participant \\
NeuroBridge \citep{Zhang2026NeuroBridge} & AAAI 2026 & Table 6, intra-subject & Within participant \\
\textbf{CORTIVA} & Current study & Main Table 3; Table~\ref{tab:supp-meg-seeds} & Within participant \\
\end{longtable}
\tableNoteSep
\noindent $^{\dagger}$ATM-S participant entries use the ten-participant results reported by UBP and NeuroBridge; the Mean column uses the ATM-S aggregate. CORTIVA uses 160 target-participant adaptation concepts and 40 validation concepts; test concepts are disjoint.
\endgroup

\Needspace{8\baselineskip}
\section{Dataset and acquisition details}

The EEG dataset is described in the THINGS-EEG2 publication \citep{Gifford2022THINGSEEG2} and publicly distributed via OSF and NeMAR \citep{Gifford2021THINGSOSF,Gifford2026THINGSEEG2NeMAR}. THINGS-data describes the MEG collection \citep{Hebart2023THINGSData}. Table~\ref{tab:supp-dataset-provenance} lists identifiers and acquisition metadata \citep{Gorgolewski2016BIDS,Pernet2019EEGBIDS}.

\begingroup
\suppTableFormat
\begin{longtable}{@{}L{0.28\textwidth}L{0.32\textwidth}L{0.32\textwidth}@{}}
\caption{Datasets and acquisition parameters used in the CORTIVA analysis.}\label{tab:supp-dataset-provenance}\\
\toprule
\tblhead{Property} & \tblhead{Value} & \tblhead{Source} \\
\midrule
\endfirsthead
\caption[]{Datasets and acquisition parameters used in the CORTIVA analysis (Continued).}\\
\toprule
\tblhead{Property} & \tblhead{Value} & \tblhead{Source} \\
\midrule
\endhead
\bottomrule
\endlastfoot
Dataset & THINGS-EEG2 & Dataset paper and NeMAR \\
NeMAR identifier & \path{NM000232} & NeMAR \\
Original OSF DOI & \path{10.17605/OSF.IO/3JK45} & OSF \\
NeMAR DOI & \path{10.82901/nemar.nm000232} & NeMAR \\
EEG participants & 10 & Dataset paper and NeMAR \\
EEG channels & 63 & Dataset paper and NeMAR \\
Acquisition sampling rate & 1000~Hz & Dataset paper and NeMAR \\
EEG analysis grid & 0--500~ms; 500~Hz; 250 samples & Current preprocessing pipeline \\
EEG preprocessing & Baseline correction, \(-100\) to 0~ms; adjacent-sample averaging; no additional filtering or artifact rejection & Current preprocessing pipeline \\
BIDS version & 1.9.0 & NeMAR \\
EEG release license & CC BY 4.0 & OSF and NeMAR \\
Primary EEG retrieval test & 200 images; four session-level test means per image are averaged & Current evaluation protocol \\
\midrule
Dataset & THINGS-MEG & THINGS-data publication \\
OpenNeuro identifier & \path{ds004212} & OpenNeuro / THINGS-data \\
MEG participants & 4 (distinct from the EEG cohort) & THINGS-data publication \\
MEG sensors & 271 & THINGS-data publication \\
MEG acquisition sampling rate & 1200~Hz & THINGS-data publication \\
MEG analysis grid & 0--1000~ms; 200~Hz & Current evaluation protocol \\
Training set & 1,654 concepts, 12 images per concept & THINGS-data publication \\
Retrieval test & 200 images, 12 repetitions per image & THINGS-data publication \\
\end{longtable}
\endgroup

\section{Ethics and data use}

CORTIVA is a secondary analysis of THINGS-EEG2 and THINGS-MEG. The dataset publications report ethics approval and consent for the original collections \citep{Gifford2022THINGSEEG2,Hebart2023THINGSData}; OSF and NeMAR record the public EEG-release license \citep{Gifford2021THINGSOSF,Gifford2026THINGSEEG2NeMAR}.

\section{Retrieval configuration and model selection}

The data split and principal hyperparameters for CORTIVA's shared SAR/CVR EEG encoder and independent EPR encoder are summarized in Table~\ref{tab:supp-endpoint-config}. Checkpoint and fusion parameters are chosen on concept-disjoint validation data before the 200-image test set is scored.

\begingroup
\suppTableFormat
\renewcommand{\arraystretch}{1.05}
\begin{longtable}{L{0.25\textwidth}L{0.24\textwidth}L{0.43\textwidth}}
\caption{CORTIVA retrieval configuration and model-selection summary.}\label{tab:supp-endpoint-config}\\
\toprule
\tblhead{Item} & \tblhead{Value} & \tblhead{Interpretation} \\
\midrule
\endfirsthead
\caption[]{CORTIVA retrieval configuration and model-selection summary (Continued).}\\
\toprule
\tblhead{Item} & \tblhead{Value} & \tblhead{Interpretation} \\
\midrule
\endhead
\bottomrule
\endlastfoot
Participants & 10 & Separate within-participant models for EEG Sub01--Sub10. \\
Training split & 1,454 concepts / 14,540 EEG inputs per participant & Ten images per concept are used for parameter estimation. \\
Concept-disjoint validation set & 200 concepts / 2,000 EEG trials per participant & Ten images per concept are aggregated for checkpoint and fusion selection. \\
Held-out retrieval test & 200 image queries and candidates; four session-level test means averaged per query & Fixed THINGS-EEG2 query and candidate bank used for final Top-1 and Top-5 scoring. \\
Training schedule & Maximum 60 epochs, patience 5 & Validation selects a checkpoint within the 60-epoch cap. \\
Optimization & AdamW \citep{Loshchilov2019AdamW}; \(\beta=(0.9,0.999)\); batch 384; eval batch 200; base LR \(2\times10^{-5}\); head/router LR \(2\times10^{-4}\); weight decay 0.01; gradient norm 1.0 & Optimizer hyperparameters. \\
Trainable modules & SAR/EPR projection and feature-normalization blocks; SAR image projector; CVR/EPR residual heads; router; route logit scales & 13.79~M updated parameters per participant; earlier backbone blocks and all target/reference networks remain frozen. \\
EEG encoders & Shared EEG encoder for SAR/CVR and independent ensemble encoder for EPR & Backbone architectures. \\
EPR spectral module & Five masks with thresholds 0.5--4, 4--8, 8--13, 13--30, and 30--45 on the module's fixed 250-Hz internal grid & Learns channel and band weights, applies an inverse FFT, and blends the result residually with the input. \\
Residual heads & CVR and EPR: \(1024\rightarrow1024\rightarrow1024\), GELU, dropout 0.1, zero-initialized final layer & Residual adaptation initialized as the identity for both auxiliary routes. \\
Router & Concatenate three 1,024-d route vectors; LayerNorm; \(3072\rightarrow512\rightarrow3\); GELU; dropout 0.05; prior \((0.55,0.25,0.20)\) & Query-conditioned route weights before score-margin confidence weighting. \\
EEG input & 63 channels; 0--500~ms; 500~Hz; 250 samples & Training session means are averaged by image; four test-session means are averaged for the primary evaluation. \\
Checkpoint selection & Concept-disjoint validation-set total loss & Selection criterion before scoring the 200-image test set. \\
Score-margin confidence selection & \(\alpha\) grid: 0, 0.1, 0.2, 0.25, 0.3, 0.4; temperature grid: 2.5, 5.0, 10.0 & Selected by validation mean rank. \\
Candidate-label permutation & 10,000 candidate-column permutations per participant and seed & Re-scoring of each five-seed score matrix after candidate identities are permuted. \\
\end{longtable}
\endgroup

\Needspace{8\baselineskip}
\section{Computational footprint}

Table~\ref{tab:supp-computational-footprint} distinguishes parameter adaptation from online routing and candidate scoring. Static counts are derived from tensor dimensions and the evaluation configuration. Measured rows use the frozen Sub01 seed-2026 model on an NVIDIA RTX 4080 SUPER with candidate banks resident; offline visual feature extraction is excluded. The multiply-accumulate subtotal covers the router's dense layers and the three candidate-bank dot products.

\begingroup
\suppDenseTableFormat
\renewcommand{\arraystretch}{1.08}
\begin{longtable}{@{}L{0.20\textwidth}L{0.23\textwidth}L{0.34\textwidth}L{0.15\textwidth}@{}}
\caption{Computational and resource footprint of within-participant EEG retrieval.}\label{tab:supp-computational-footprint}\\
\toprule
\tblhead{Resource or operation} & \tblhead{Quantity} & \tblhead{Derivation and scope} & \tblhead{Stage} \\
\midrule
\endfirsthead
\caption[]{Computational and resource footprint of within-participant EEG retrieval (Continued).}\\
\toprule
\tblhead{Resource or operation} & \tblhead{Quantity} & \tblhead{Derivation and scope} & \tblhead{Stage} \\
\midrule
\endhead
\bottomrule
\endlastfoot
Trainable parameters & 13.79~M (13,791,777) per participant & Final encoder blocks, prediction heads, and router; frozen visual targets remain fixed. & Training \\
Query router & 1,581,059 parameters; 1,574,400 dense multiply-accumulates per query & LayerNorm over 3,072 features, \(3072\rightarrow512\rightarrow3\) dense layers. & Training and online \\
Inference candidate banks & \(3\times200\times1024=614{,}400\) float32 values (2.344~MiB) & One final 1,024-d candidate matrix per route after model selection. & Online \\
Visual feature sources & Excluded from the 13.8-million trainable count & Source feature extraction is offline; the selected SAR projector generates its final bank, while CVR and EPR load fixed targets. & Offline / finalization \\
Candidate dot products & 614,400 multiply-accumulates per query & Three routes, 200 candidates, and 1,024 dimensions; excludes normalization and the final weighted sum. & Online \\
Measured inference, batch 1 & 6.30~ms median (IQR 0.22~ms); 92.26~MiB peak allocated (7.83~MiB above the loaded-model baseline) & Device-resident end-to-end inference over 1,000 timed iterations after 100 warm-up iterations. & Online \\
Measured inference, batch 200 & 14.22~ms median (IQR 0.12~ms); 663.83~MiB peak allocated (564.65~MiB incremental) & Device-resident end-to-end inference over 300 timed iterations after 100 warm-up iterations. & Online \\
Training input & \(14{,}540\times63\times250\) float32 values (873.59~MiB) & Uncompressed participant-specific training tensor. & Training \\
Validation input & \(2{,}000\times63\times250\) float32 values (120.16~MiB) & Concept-disjoint tensor used for checkpoint and fusion selection. & Validation \\
Optimization budget & 38 batches per epoch; at most 2,280 updates per participant & \(\lceil14{,}540/384\rceil\times60\), with early-stopping patience 5. & Training \\
Confidence search & \(6\times3=18\) configurations per participant & Six blend weights and three confidence temperatures evaluated on validation data. & Validation \\
\end{longtable}
\endgroup

\Needspace{8\baselineskip}
\section{THINGS-MEG configuration and seed stability}

The MEG evaluation uses the SAR, CVR, and EPR target families with candidate-score fusion. Its model uses 512-dimensional route spaces, a 271-sensor stem, and a modality-specific neural encoder. Table~\ref{tab:supp-meg-config} gives the training and selection configuration.

\begingroup
\suppTableFormat
\renewcommand{\arraystretch}{1.05}
\begin{longtable}{@{}L{0.25\textwidth}L{0.26\textwidth}L{0.41\textwidth}@{}}
\caption{THINGS-MEG CORTIVA configuration.}\label{tab:supp-meg-config}\\
\toprule
\tblhead{Item} & \tblhead{Value} & \tblhead{Role} \\
\midrule
\endfirsthead
\caption[]{THINGS-MEG CORTIVA configuration (Continued).}\\
\toprule
\tblhead{Item} & \tblhead{Value} & \tblhead{Role} \\
\midrule
\endhead
\bottomrule
\endlastfoot
Participants & Sub01--Sub04 (MEG cohort) & Separate within-participant models; labels are distinct from the EEG cohort. \\
MEG input & 271 sensors; 0--1~s; 200~Hz; participant-wise standardized & Input to the modality-specific sensor and temporal stem. \\
Training set & 1,654 concepts; 12 images per concept & Participant-specific parameter estimation and full-data refit. \\
Selection split & 1,454 training concepts / 200 concept-disjoint validation concepts & Selects the checkpoint and score-margin confidence parameters. \\
Test retrieval & 200 image queries and candidates; 12 MEG repetitions averaged per query & Final Top-1 and Top-5 scoring. \\
Routes & SAR, CVR, and EPR; 512-d projected score spaces & Three target-specific candidate-score vectors. \\
Neural encoders & EEGProject for SAR/CVR; TSConv for EPR \citep{Zhang2026NeuroBridge} & Modality-specific feature extraction. \\
SAR warm-up & Maximum 15 epochs & Selects the SAR initialization on validation mean rank. \\
Joint optimization & Maximum 60 epochs; patience 12; batch 1,024 & Trains auxiliary routes and query-conditioned router. \\
Optimizer & AdamW; learning rate \(10^{-4}\); weight decay \(10^{-4}\) & Shared optimization setting across all six seeds. \\
Training augmentation & Five-sample temporal smoothing independently applied to each channel with probability 0.30 & Applied stochastically during training. \\
Confidence selection & \(\alpha\): 0, 0.1, 0.2, 0.25, 0.3, 0.4; \(T_c\): 2.5, 5, 10 & Validation-selected score-margin confidence weighting. \\
Full-data refit & Selected warm-up and joint durations on all 1,654 training concepts & Produces the final checkpoint before test scoring. \\
Optimization seeds & 2026--2031 & Six independent training runs. \\
\end{longtable}
\endgroup

\Needspace{16\baselineskip}
\begingroup
\suppTableFormat
\begin{longtable}{@{}L{0.20\textwidth}C{0.22\textwidth}C{0.22\textwidth}@{}}
\caption{Six-seed THINGS-MEG retrieval and participant-level stability. Sub01--Sub04 denote the MEG cohort and are distinct from the identically numbered EEG participants.}\label{tab:supp-meg-seeds}\\
\toprule
\multicolumn{3}{c}{\tblhead{A. Seed-level four-participant means}} \\
\cmidrule(lr){1-3}
\tblhead{Seed} & \tblhead{Top-1 (\%)} & \tblhead{Top-5 (\%)} \\
\midrule
\endfirsthead
\caption[]{Six-seed THINGS-MEG retrieval and participant-level stability (Continued).}\\
\toprule
\tblhead{Summary} & \tblhead{Top-1 (\%)} & \tblhead{Top-5 (\%)} \\
\midrule
\endhead
\bottomrule
\endlastfoot
2026 & 43.000 & 73.750 \\
2027 & 42.125 & 73.000 \\
2028 & 43.375 & 75.000 \\
2029 & 40.875 & 71.875 \\
2030 & 40.625 & 73.875 \\
2031 & 44.500 & 74.375 \\
\midrule
Mean $\pm$ SD & $42.417\pm1.501$ & $73.646\pm1.094$ \\
\addlinespace[5pt]
\multicolumn{3}{c}{\tblhead{B. Participant-level summaries across six seeds}} \\
\cmidrule(lr){1-3}
\tblhead{Participant} & \tblhead{Top-1 (\%)} & \tblhead{Top-5 (\%)} \\
\midrule
Sub01 & $20.583\pm0.801$ & $50.750\pm2.019$ \\
Sub02 & $65.583\pm4.421$ & $91.250\pm1.891$ \\
Sub03 & $52.583\pm2.990$ & $86.500\pm1.871$ \\
Sub04 & $30.917\pm1.201$ & $66.083\pm1.625$ \\
\end{longtable}
\endgroup

\Needspace{8\baselineskip}
\section{Route and visual-target specifications}

Each EEG route is trained against a target candidate geometry derived from one or more frozen visual encoders. CLIP-RN50 supplies both SAR and EPR features \citep{Radford2021CLIP}; OpenCLIP ViT-B/32, SynCLR, and SDXL-VAE form the composite visual sources for CVR \citep{Ilharco2021OpenCLIP,Cherti2023OpenCLIP,Schuhmann2022LAION5B,Tian2024SynCLR,Kingma2014VAE,Podell2023SDXL}. SAR and EPR share a feature source but use different target transforms: SAR updates its image projector during training, whereas EPR averages three frozen projections.

\begingroup
\suppTableFormat
\begin{longtable}{L{0.16\textwidth}L{0.23\textwidth}L{0.31\textwidth}L{0.22\textwidth}}
\caption{CORTIVA route-to-target mapping.}\label{tab:supp-route-map}\\
\toprule
\tblhead{Route} & \tblhead{Neural encoder} & \tblhead{Visual source and target construction} & \tblhead{Online role} \\
\midrule
\endfirsthead
\caption[]{CORTIVA route-to-target mapping (Continued).}\\
\toprule
\tblhead{Route} & \tblhead{Neural encoder} & \tblhead{Visual source and target construction} & \tblhead{Online role} \\
\midrule
\endhead
\bottomrule
\endlastfoot
Semantic Alignment Route (SAR) & Shared graph-token EEG encoder; 1,024-d output & CLIP-RN50 features followed by the trainable SAR image projector & Produces \(S^{(\mathrm{sar})}\) over the 200 candidates. \\
Composite Visual Route (CVR) & Residual CVR head applied to the SAR EEG vector & OpenCLIP-B/32, SynCLR, and SDXL-VAE features transformed by fixed source-specific projectors and fused by a preceding composite-target model into a normalized 1,024-d target & Produces \(S^{(\mathrm{cvr})}\). \\
Ensemble Projection Route (EPR) & Independent spectral--temporal encoder plus residual EPR head; 1,024-d output & Normalized mean of three frozen image-projection checkpoints applied to CLIP-RN50 features, followed by final normalization & Produces \(S^{(\mathrm{epr})}\). \\
\textbf{Final CORTIVA score} & Query-conditioned router blended with validation-selected score-margin weights & Weighted sum of the three route-specific 200-way score vectors & Produces \(S^{(\mathrm{fused})}\). \\
\end{longtable}
\endgroup

Stage~1 performs participant-specific, single-target EEG--image alignment on the training and validation concepts. It produces the paired EEG and image-projection checkpoints used to initialize Stage~2, including the CVR source projectors and composite-target model and the three EPR projection pairs. Stage~2 freezes those target/reference networks and trains CORTIVA for 60 epochs. Participant-specific initialization pairs are selected by Stage~1 validation loss.

Candidate scoring for route \(b\) uses the temperature-scaled cosine similarity in Eq.~(1) of the main text. The router receives concatenated SAR, CVR, and EPR EEG vectors. For each route, let \(S^{(b)}_{si(1)}\geq S^{(b)}_{si(2)}\) be the two largest candidate scores for query \(i\). The score-margin confidence vector is
\begin{equation}
\begin{aligned}
m^{(b)}_{s,i} &= S^{(b)}_{si(1)}-S^{(b)}_{si(2)}, \\
\mu_{s,i} &= |\mathcal{B}|^{-1}\sum_{b\in\mathcal{B}}m^{(b)}_{s,i}, \\
\sigma_{s,i} &= \left[|\mathcal{B}|^{-1}\sum_{b\in\mathcal{B}}\left(m^{(b)}_{s,i}-\mu_{s,i}\right)^2\right]^{1/2}, \\
\widehat m^{(b)}_{s,i} &= \frac{m^{(b)}_{s,i}-\mu_{s,i}}{\sigma_{s,i}+10^{-6}}, \\
c^{(b)}_{s,i} &= \frac{\exp\!\left(\widehat m^{(b)}_{s,i}/T_{c,s}\right)}{\sum_{b'\in\mathcal{B}}\exp\!\left(\widehat m^{(b')}_{s,i}/T_{c,s}\right)}.
\end{aligned}
\label{eq:supp-margin-confidence}
\end{equation}
Participant-specific \(\alpha_s\) and \(T_{c,s}\) blend \(c_{s,i}\) with the learned router weights before ranking. Candidate-score fusion preserves the learned inverse-temperature scale of each route vector.

\section{Training objective and validation-selected fusion parameters}
\label{sec:supp-objective}

The training objective contains 15 terms whose weights are fixed globally across participants, seeds, and matched ablations. Let \(\bar{x}=x/\lVert x\rVert_2\), let \(I_B=(1,\ldots,B)\) denote diagonal labels for a batch of size \(B\), and let softmax operate row-wise. The symmetric contrastive, cosine-preservation, and relational losses are
\begin{gather}
\mathcal{C}_{\tau_b}(A,V)
=\tfrac{1}{2}\!\left[\operatorname{CE}(\tau_b\bar A\bar V^{\mathsf T},I_B)
+\operatorname{CE}(\tau_b\bar V\bar A^{\mathsf T},I_B)\right], \\
\mathcal{D}_{\mathrm{cos}}(A,T)
=1-\frac{1}{B}\sum_{i=1}^{B}\bar a_i^{\mathsf T}\bar t_i, \\
P_{\tau}(A)
=\operatorname{softmax}(\bar A\bar A^{\mathsf T}/\tau), \\
\mathcal{D}_{\mathrm{rel}}(A,T)
=\tau^2\operatorname{KL}\!\left(P_{\tau}(T)\,\|\,P_{\tau}(A)\right).
\end{gather}
where \(\tau_b=\min[\exp(\eta_b),100]\) is the learned inverse temperature of route \(b\), matching the main text. For candidate-distribution distillation, define
\begin{gather}
P_{\tau}(A;V)
 =\operatorname{softmax}(\bar A\bar V^{\mathsf T}/\tau), \\
\mathcal{D}_{\mathrm{logit}}(A,T;V)
 =\tau^2\operatorname{KL}\!\left(P_{\tau}(T;V)\,\|\,P_{\tau}(A;V)\right).
\end{gather}
For the \(M\)-member EPR projection ensemble, the target distribution is the arithmetic mean of member probabilities:
\begin{gather}
P^{\mathrm{ens}}_{A}
 =\frac{1}{M}\sum_{m=1}^{M}P_{\tau}(A;V_m), \\
P^{\mathrm{ens}}_{T}
 =\frac{1}{M}\sum_{m=1}^{M}P_{\tau}(T_m;V_m), \\
\mathcal{D}_{\mathrm{ens}}(A,\{T_m,V_m\})
 =\tau^2\operatorname{KL}\!\left(P^{\mathrm{ens}}_{T}\,\|\,P^{\mathrm{ens}}_{A}\right).
\end{gather}
Every KL divergence is averaged over batch rows.

\Needspace{8\baselineskip}
Let \(Z_{\mathrm{s}}\) and \(V_{\mathrm{s}}\) be the SAR EEG and trainable projected-image representations, \(X_{\mathrm{t}}\) the frozen text representation, and \(\widetilde Z_{\mathrm{s}},\widetilde V_{\mathrm{s}}\) the normalized SAR reference consensus. Let \(Z_{\mathrm{c}},V_{\mathrm{c}}\) denote the CVR prediction and frozen composite target. Let \(Z_{\mathrm{e}}\) be the EPR prediction, \(\widetilde Z_{\mathrm{e}}\) and \(V_{\mathrm{e}}\) the normalized means of the frozen EPR EEG and image reference projections, and \(\{\widetilde Z_{\mathrm{e},r},V_{\mathrm{e},r}\}_{r=1}^{M}\) their individual pairs. The objective is
\begin{equation}
\begin{aligned}
\mathcal{L}={}&
0.20\,\mathcal{C}_{\tau_{\mathrm{s}}}(Z_{\mathrm{s}},V_{\mathrm{s}})
+0.03\,\mathcal{C}_{\tau_{\mathrm{s}}}(V_{\mathrm{s}},X_{\mathrm{t}})
+0.08\,\mathcal{D}_{\mathrm{cos}}(Z_{\mathrm{s}},\widetilde Z_{\mathrm{s}}) \\
&+0.03\,\mathcal{D}_{\mathrm{rel}}(Z_{\mathrm{s}},\widetilde Z_{\mathrm{s}})
+0.03\,\mathcal{D}_{\mathrm{logit}}(Z_{\mathrm{s}},\widetilde Z_{\mathrm{s}};\widetilde V_{\mathrm{s}})
+0.02\,\mathcal{D}_{\mathrm{cos}}(V_{\mathrm{s}},\widetilde V_{\mathrm{s}}) \\
&+0.18\,\mathcal{C}_{\tau_{\mathrm{c}}}(Z_{\mathrm{c}},V_{\mathrm{c}})
+0.05\,\mathcal{D}_{\mathrm{cos}}(Z_{\mathrm{c}},V_{\mathrm{c}})
+0.04\,\mathcal{D}_{\mathrm{rel}}(Z_{\mathrm{c}},V_{\mathrm{c}}) \\
&+0.18\,\mathcal{C}_{\tau_{\mathrm{e}}}(Z_{\mathrm{e}},V_{\mathrm{e}})
+0.05\,\mathcal{D}_{\mathrm{cos}}(Z_{\mathrm{e}},\widetilde Z_{\mathrm{e}})
+0.04\,\mathcal{D}_{\mathrm{rel}}(Z_{\mathrm{e}},\widetilde Z_{\mathrm{e}}) \\
&+0.04\,\mathcal{D}_{\mathrm{ens}}(Z_{\mathrm{e}},\{\widetilde Z_{\mathrm{e},r},V_{\mathrm{e},r}\}_{r=1}^{M})
+0.05\,\mathcal{L}_{\mathrm{rce}}
+0.04\,\mathcal{L}_{\mathrm{trace}}.
\end{aligned}
\label{eq:supp-full-objective}
\end{equation}
The relational and candidate-distribution temperatures are both \(\tau=0.07\), and the router-trace target temperature is 5. Table~\ref{tab:supp-losses} lists the 15 terms and their fixed weights. Training and evaluation use PyTorch \citep{Paszke2019PyTorch} and NumPy \citep{Harris2020NumPy}.

\begingroup
\suppTableFormat
\begin{longtable}{L{0.26\textwidth}C{0.09\textwidth}L{0.57\textwidth}}
\caption{Training-loss terms in CORTIVA.}\label{tab:supp-losses}\\
\toprule
\tblhead{Loss term} & \tblhead{Weight} & \tblhead{Function} \\
\midrule
\endfirsthead
\caption[]{Training-loss terms in CORTIVA (Continued).}\\
\toprule
\tblhead{Loss term} & \tblhead{Weight} & \tblhead{Function} \\
\midrule
\endhead
\bottomrule
\endlastfoot
SAR symmetric contrastive & 0.20 & Encourages agreement between SAR EEG and projected CLIP-RN50 image representations. \\
Image--text contrastive & 0.03 & Encourages image--text agreement in the trainable semantic image projection. \\
SAR reference-direction cosine & 0.08 & Penalizes deviation from the normalized SAR reference direction. \\
SAR relational KL & 0.03 & Penalizes deviations from the SAR reference ensemble's within-batch relational geometry. \\
SAR candidate-logit KL & 0.03 & Penalizes divergence from SAR reference candidate distributions. \\
Image-projection preservation & 0.02 & Penalizes deviation from the reference semantic image projection. \\
CVR contrastive & 0.18 & Encourages agreement between the CVR residual route and the frozen composite visual target. \\
CVR cosine & 0.05 & Penalizes deviation from the normalized CVR target direction. \\
CVR relational & 0.04 & Penalizes deviations from within-batch composite-target geometry. \\
EPR contrastive & 0.18 & Encourages agreement between the independent EPR route and the frozen projection ensemble. \\
EPR reference cosine & 0.05 & Penalizes deviation from the normalized EPR reference-ensemble EEG direction. \\
EPR relational & 0.04 & Penalizes deviations from EPR reference relational geometry. \\
EPR ensemble-logit KL & 0.04 & Penalizes divergence from candidate distributions of the frozen EPR projection ensemble. \\
Routed symmetric cross-entropy & 0.05 & Trains the fused route-score matrix in both EEG-to-image and image-to-EEG directions. \\
Router trace cross-entropy & 0.04 & Encourages router weights to follow route-indexed reference-score margins on training rows. \\
\end{longtable}
\endgroup

\begingroup
\suppTableFormat
\begin{longtable}{L{0.19\textwidth}C{0.15\textwidth}C{0.15\textwidth}C{0.20\textwidth}C{0.20\textwidth}}
\caption{Five-seed validation-selected score-margin confidence parameters. Parameter columns give the range across seeds 2026--2030; validation metrics are mean $\pm$ SD across seeds.}\label{tab:supp-subject-blend}\\
\toprule
\tblhead{Participant} & \tblhead{\(\alpha_s\) range} & \tblhead{\(T_{c,s}\) range} & \tblhead{Val. Top-1 (\%)} & \tblhead{Val. mean rank} \\
\midrule
\endfirsthead
\caption[]{Five-seed validation-selected score-margin confidence parameters (Continued).}\\
\toprule
\tblhead{Participant} & \tblhead{\(\alpha_s\) range} & \tblhead{\(T_{c,s}\) range} & \tblhead{Val. Top-1 (\%)} & \tblhead{Val. mean rank} \\
\midrule
\endhead
\bottomrule
\endlastfoot
Sub01 & 0.0 & 2.5 & $99.5\pm0.0$ & $1.010\pm0.000$ \\
Sub02 & 0.0 & 2.5 & $100.0\pm0.0$ & $1.000\pm0.000$ \\
Sub03 & 0.0--0.1 & 2.5 & $98.9\pm0.7$ & $1.011\pm0.007$ \\
Sub04 & 0.0 & 2.5 & $99.4\pm0.2$ & $1.006\pm0.002$ \\
Sub05 & 0.0--0.4 & 2.5--5.0 & $98.5\pm0.5$ & $1.015\pm0.005$ \\
Sub06 & 0.0 & 2.5 & $100.0\pm0.0$ & $1.000\pm0.000$ \\
Sub07 & 0.0 & 2.5 & $100.0\pm0.0$ & $1.000\pm0.000$ \\
Sub08 & 0.0 & 2.5 & $99.5\pm0.0$ & $1.005\pm0.000$ \\
Sub09 & 0.0 & 2.5 & $99.9\pm0.2$ & $1.001\pm0.002$ \\
Sub10 & 0.0 & 2.5 & $100.0\pm0.0$ & $1.000\pm0.000$ \\
\end{longtable}
\endgroup

The 2,000 validation observations comprise 200 concepts with ten images per concept. Candidate features and score rows are averaged within concept to form the \(200\times200\) validation matrix. The final test contains one held-out image per concept and one query per candidate.

Concept-level aggregation produces validation Top-1 values of 98.5--100.0\% and mean ranks of 1.000--1.015 across participants. Eight of ten participants select \(\alpha_s=0\) for all five seeds; Sub03 and Sub05 select a nonzero confidence contribution in only a subset of seeds. Thus, the score-margin term is inactive for most EEG configurations.

\section{EEG input and index alignment}
\label{sec:supp-eeg-input}

The full input contains all 63 channels over 0--500~ms at 500~Hz, represented by 250 samples. Session-level trial means are averaged within image, producing 16,540 image-level development inputs: 14,540 for fitting and 2,000 for validation. The primary held-out evaluation averages four test-session means into one \(63\times250\) query tensor per image; the repetition analysis evaluates every combination of one through four sessions.

\begingroup
\suppTableFormat
\begin{longtable}{L{0.22\textwidth}L{0.24\textwidth}L{0.27\textwidth}L{0.17\textwidth}}
\caption{EEG input surfaces and orthogonal negative controls for 60-epoch CORTIVA retrieval.}\label{tab:supp-eeg-input-surfaces}\\
\toprule
\tblhead{Condition} & \tblhead{Input timing} & \tblhead{Channels and processing} & \tblhead{Role} \\
\midrule
\endfirsthead
\caption[]{EEG input surfaces and orthogonal negative controls for 60-epoch CORTIVA retrieval (Continued).}\\
\toprule
\tblhead{Condition} & \tblhead{Input timing} & \tblhead{Channels and processing} & \tblhead{Role} \\
\midrule
\endhead
\bottomrule
\endlastfoot
Full CORTIVA input & 0--500~ms; 500~Hz; 250 samples & All 63 channels; four-session average before scoring & Primary input \\
Candidate-label permutation & Same tensor as the full CORTIVA input & Candidate identities permuted after scoring & Indexing check \\
Pre-stimulus input & \(-500\) to 0~ms; 500~Hz; 250 samples & Real duration-matched EEG from all 63 channels & Temporal control \\
Frontal-only input & 0--500~ms; 500~Hz; 250 samples & Frontal sensors retained; all other channels set to zero & Spatial control \\
\end{longtable}
\endgroup

Together, the controls disrupt query--candidate identity, post-stimulus timing, or posterior spatial information while preserving the frozen CORTIVA endpoint and its 200-way candidate bank.

\section{Visual target specifications}

CLIP-RN50 supplies SAR and EPR features \citep{Radford2021CLIP}; OpenCLIP ViT-B/32, SynCLR, and SDXL-VAE form the composite visual target \citep{Ilharco2021OpenCLIP,Cherti2023OpenCLIP,Schuhmann2022LAION5B,Tian2024SynCLR,Kingma2014VAE,Podell2023SDXL}.

\begingroup
\suppTableFormat
\begin{longtable}{@{}L{0.34\textwidth}L{0.14\textwidth}C{0.07\textwidth}C{0.07\textwidth}L{0.28\textwidth}@{}}
\caption{Frozen visual sources used in CORTIVA. Development rows comprise 14,540 fitting and 2,000 validation images.}\label{tab:supp-visual-targets}\\
\toprule
\tblhead{Source key} & \tblhead{Layer} & \tblhead{Dev. rows} & \tblhead{Test rows} & \tblhead{Role} \\
\midrule
\endfirsthead
\caption[]{Frozen visual sources used in CORTIVA (Continued).}\\
\toprule
\tblhead{Source key} & \tblhead{Layer} & \tblhead{Dev. rows} & \tblhead{Test rows} & \tblhead{Role} \\
\midrule
\endhead
\bottomrule
\endlastfoot
\path{clip-rn50-features} & Candidate source & 16,540 & 200 & 1,024-d source for SAR and EPR candidate construction \\
\texttt{CLIP-ViT-B-32-}\allowbreak\texttt{laion2B-s34B-b79K} & Composite-target input & 16,540 & 200 & OpenCLIP ViT-B/32, LAION-2B-en, 512-d \\
\path{synclr_vit_b_16} & Composite-target input & 16,540 & 200 & SynCLR ViT-B/16, 768-d \\
\path{SDXL-VAE} & Composite-target input & 16,540 & 200 & SDXL AutoencoderKL; \(16\times16\times4\) latent flattened to 1,024-d \\
\end{longtable}
\endgroup

\section{Retrieval summary}

SD and range quantify between-participant variation, whereas intervals quantify uncertainty in the group means.

\begingroup
\suppDenseTableFormat
\begin{longtable}{L{0.28\textwidth}C{0.07\textwidth}C{0.10\textwidth}C{0.10\textwidth}C{0.16\textwidth}C{0.17\textwidth}}
\caption{CORTIVA aggregate statistics over participant-level summaries.}\label{tab:supp-cortiva-aggregate-statistics}\\
\toprule
\tblhead{Metric} & \tblhead{\(n\)} & \tblhead{Mean} & \tblhead{SD} & \tblhead{95\% CI} & \tblhead{Min--Max} \\
\midrule
\endfirsthead
\caption[]{CORTIVA aggregate statistics over participant-level summaries (Continued).}\\
\toprule
\tblhead{Metric} & \tblhead{\(n\)} & \tblhead{Mean} & \tblhead{SD} & \tblhead{95\% CI} & \tblhead{Min--Max} \\
\midrule
\endhead
\bottomrule
\endlastfoot
Top-1 retrieval (\%) & 10 & 73.53 & 6.59 & 69.62--77.33 & 63.90--81.50 \\
Top-5 retrieval (\%) & 10 & 95.31 & 2.37 & 93.81--96.56 & 89.80--97.60 \\
Mean rank (candidates) & 10 & 1.842 & 0.372 & 1.654--2.083 & 1.493--2.749 \\
\end{longtable}
\endgroup

\begingroup
\suppTableFormat
\begin{longtable}{L{0.22\textwidth}L{0.24\textwidth}C{0.13\textwidth}C{0.13\textwidth}C{0.14\textwidth}}
\caption{Orthogonal negative-control summary for 60-epoch CORTIVA retrieval. Values are means over ten participant-level five-seed summaries; candidate-label values additionally average 10,000 permutations per participant and seed.}\label{tab:supp-cortiva-controls}\\
\toprule
\tblhead{Condition} & \tblhead{Information disrupted} & \tblhead{Top-1 (\%)} & \tblhead{Top-5 (\%)} & \tblhead{Mean rank} \\
\midrule
\endfirsthead
\caption[]{Orthogonal negative-control summary for 60-epoch CORTIVA retrieval (Continued).}\\
\toprule
\tblhead{Condition} & \tblhead{Information disrupted} & \tblhead{Top-1 (\%)} & \tblhead{Top-5 (\%)} & \tblhead{Mean rank} \\
\midrule
\endhead
\bottomrule
\endlastfoot
\textbf{CORTIVA} & None & \textbf{73.53} & \textbf{95.31} & \textbf{1.842} \\
Candidate-label permutation & Query--candidate identity & 0.500 & 2.500 & 100.493 \\
Pre-stimulus input & Post-stimulus timing & 0.60 & 2.41 & 100.069 \\
Frontal-only input & Posterior sensor information & 0.54 & 2.41 & 99.831 \\
Analytical chance & --- & 0.50 & 2.50 & 100.5 \\
\end{longtable}
\endgroup

\begin{figure}[!t]
\centering
\includegraphics[width=0.96\linewidth]{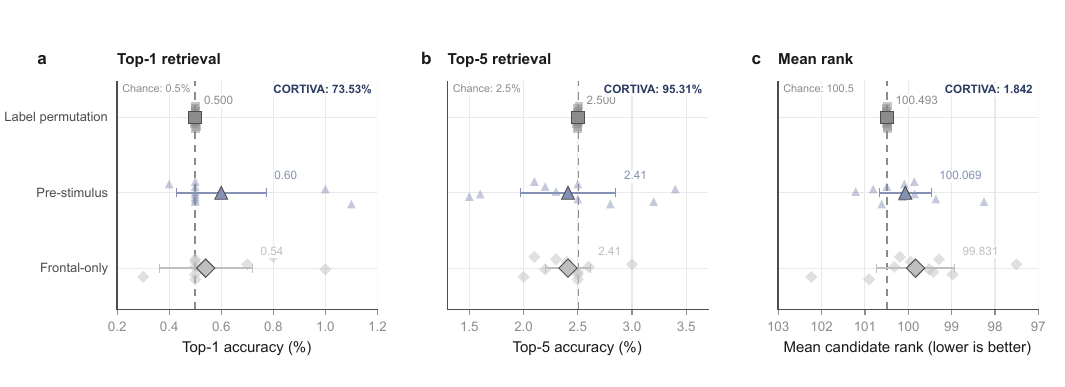}
\caption{\textbf{Three orthogonal negative controls at the analytical-chance scale.} \textbf{a--c}, Participant-level Top-1, Top-5, and mean-rank results for candidate-label permutation, duration-matched pre-stimulus EEG, and frontal-only post-stimulus input. Small markers denote participant means after five-seed averaging; large markers and whiskers denote group means and 95\% \(t\)-intervals. Dashed lines mark analytical chance, and navy annotations give the corresponding CORTIVA endpoints. The controls isolate query--candidate identity, post-stimulus timing, and posterior spatial information. Holm-adjusted exact sign-flip tests separate CORTIVA from candidate-label permutation across all three metrics (\(p_{\mathrm{Holm}}=0.0059\)).}
\label{fig:supp-e60-negative-controls}
\end{figure}

\FloatBarrier
\Needspace{28\baselineskip}
\section{CORTIVA component comparison}

\begingroup
\suppDenseTableFormat
\renewcommand{\arraystretch}{1.04}
\begin{longtable}{L{0.22\textwidth}L{0.10\textwidth}*{2}{C{0.085\textwidth}}*{2}{C{0.075\textwidth}}L{0.24\textwidth}}
\caption{CORTIVA component comparison. Values are ten-participant means on the same 200-way test; deltas are CORTIVA minus the row value.}\label{tab:supp-component-summary}\\
\toprule
\tblhead{Configuration} & \tblhead{Family} & \tblhead{Top-1 (\%)} & \tblhead{Top-5 (\%)} & \tblhead{\(\Delta\)T1 (pp)} & \tblhead{\(\Delta\)T5 (pp)} & \tblhead{Definition} \\
\midrule
\endfirsthead
\caption[]{CORTIVA component comparison (Continued).}\\
\toprule
\tblhead{Configuration} & \tblhead{Family} & \tblhead{Top-1 (\%)} & \tblhead{Top-5 (\%)} & \tblhead{\(\Delta\)T1 (pp)} & \tblhead{\(\Delta\)T5 (pp)} & \tblhead{Definition} \\
\midrule
\endhead
\bottomrule
\endlastfoot
Semantic Alignment Route (SAR) & route output & 63.81 & 90.48 & 9.72 & 4.83 & Graph-token EEG encoder scored against projected CLIP-RN50 candidates. \\
Composite Visual Route (CVR) & route output & 57.54 & 88.19 & 15.99 & 7.12 & Residual CVR head scored against the frozen composite visual target. \\
Frozen composite visual target & reference diagnostic & 49.55 & 81.00 & 23.98 & 14.31 & Frozen target-side output scored against its matching visual target. \\
Ensemble Projection Route (EPR) & route output & 62.63 & 89.45 & 10.90 & 5.86 & Independent EPR encoder scored against its image-projection ensemble. \\
Frozen EPR projection ensemble & reference diagnostic & 65.00 & 91.45 & 8.53 & 3.86 & Frozen projection-ensemble scores. \\
Fixed-weight fusion & fusion variant & 70.57 & 94.56 & 2.96 & 0.75 & Route fusion with a prespecified fixed weight vector. \\
Validation-rank fusion & fusion variant & 70.84 & 94.61 & 2.69 & 0.70 & Query-weighted negative-rank fusion before final candidate-score fusion. \\
\textbf{CORTIVA} & \textbf{fused model} & \textbf{73.53} & \textbf{95.31} & \textbf{0.00} & \textbf{0.00} & Validation-selected fusion of three 200-way score vectors. \\
\end{longtable}
\endgroup

Relative to negative-rank fusion, CORTIVA gains 2.69 pp Top-1 and 0.70 pp Top-5. The route-weight controls below test whether this advantage depends on query-aligned weights.

\clearpage
\section{Matched structural ablations}

Each leave-one-route-out model is independently retrained under the matched design used for full CORTIVA. Participant metrics average five seeds; losses are full minus ablated for accuracy and ablated minus full for mean rank.

\begingroup
\suppDenseTableFormat
\begin{longtable}{@{}L{0.12\textwidth}L{0.175\textwidth}C{0.06\textwidth}L{0.175\textwidth}C{0.06\textwidth}L{0.235\textwidth}C{0.06\textwidth}@{}}
\caption{Five-seed 60-epoch structural ablation statistics. Each interval is a simultaneous 95\% participant-bootstrap interval within the corresponding metric; \(p\)-values are exact two-sided sign-flip tests Holm-adjusted across all nine route-by-metric tests. Top-1 paired standardized effects \(d_z\) are 1.04, 1.88, and 1.42 for SAR, CVR, and EPR removal.}\label{tab:supp-structural-ablation}\\
\toprule
\tblhead{Removed route} & \tblhead{Top-1 loss [95\% CI] (pp)} & \tblhead{(p)} & \tblhead{Top-5 loss [95\% CI] (pp)} & \tblhead{(p)} & \tblhead{Mean-rank loss [95\% CI]} & \tblhead{(p)} \\
\midrule
\endfirsthead
\caption[]{Five-seed 60-epoch structural ablation statistics (Continued).}\\
\toprule
\tblhead{Removed route} & \tblhead{Top-1 loss [95\% CI] (pp)} & \tblhead{(p)} & \tblhead{Top-5 loss [95\% CI] (pp)} & \tblhead{(p)} & \tblhead{Mean-rank loss [95\% CI]} & \tblhead{(p)} \\
\midrule
\endhead
\bottomrule
\endlastfoot
SAR & (0.95\;[-1.31,3.21]) & 0.055 & (0.70\;[-0.47,1.87]) & 0.055 & (0.074\;[-0.177,0.326]) & 0.055 \\
CVR & (5.47\;[3.21,7.73]) & 0.018 & (3.03\;[1.86,4.20]) & 0.023 & (0.585\;[0.334,0.837]) & 0.018 \\
EPR & (5.40\;[3.14,7.66]) & 0.018 & (1.26\;[0.09,2.43]) & 0.055 & (0.257\;[0.005,0.508]) & 0.029 \\
\end{longtable}
\endgroup

CVR and EPR show robust removal costs, with simultaneous Top-1 intervals of 3.21--7.73 and 3.14--7.66 pp, respectively. SAR removal yields 0.95 pp with a simultaneous 95\% interval of \(-1.31\) to 3.21 pp.

\section{Route-weight controls}
\label{sec:supp-router-controls}

Table~\ref{tab:supp-router-controls} quantifies the sensitivity of query-aligned weights to four controls across the 50 route-score matrices. Accuracy effects are query-aligned minus control accuracy, and mean-rank effects are control minus query-aligned rank. The shuffled control averages 2,000 within-matrix permutations per participant. Intervals use 50,000 participant bootstrap resamples, and exact sign-flip tests are Holm-adjusted within each metric.

\begingroup
\suppDenseTableFormat
\begin{longtable}{@{}L{0.19\textwidth}L{0.10\textwidth}C{0.11\textwidth}C{0.11\textwidth}C{0.27\textwidth}C{0.10\textwidth}@{}}
\caption{Paired route-weight controls. Effects compare query-aligned weights with each control after within-participant seed averaging. Values are means with 95\% participant-bootstrap intervals; positive effects favor query alignment.}\label{tab:supp-router-controls}\\
\toprule
\tblhead{Weight control} & \tblhead{Metric} & \tblhead{Aligned} & \tblhead{Control} & \tblhead{Effect [95\% CI]} & \tblhead{\(p_{\mathrm{Holm}}\)} \\
\midrule
\endfirsthead
\caption[]{Paired route-weight controls (Continued).}\\
\toprule
\tblhead{Weight control} & \tblhead{Metric} & \tblhead{Aligned} & \tblhead{Control} & \tblhead{Effect [95\% CI]} & \tblhead{\(p_{\mathrm{Holm}}\)} \\
\midrule
\endhead
\bottomrule
\endlastfoot
Shuffled within participant & Top-1 (\%) & 73.53 & 73.42 & \(0.11\;[-0.08,0.31]\) & 0.70 \\
& Top-5 (\%) & 95.31 & 95.42 & \(-0.11\;[-0.33,0.15]\) & 0.96 \\
& Mean rank & 1.842 & 1.844 & \(0.002\;[-0.012,0.020]\) & 1.00 \\
\addlinespace[1.5pt]
Participant-mean weights & Top-1 (\%) & 73.53 & 73.76 & \(-0.23\;[-0.47,0.01]\) & 0.47 \\
& Top-5 (\%) & 95.31 & 95.63 & \(-0.32\;[-0.58,-0.04]\) & 0.27 \\
& Mean rank & 1.842 & 1.822 & \(-0.020\;[-0.035,-0.002]\) & 0.23 \\
\addlinespace[1.5pt]
Uniform weights & Top-1 (\%) & 73.53 & 74.22 & \(-0.69\;[-1.37,0.09]\) & 0.47 \\
& Top-5 (\%) & 95.31 & 95.49 & \(-0.18\;[-0.63,0.25]\) & 0.96 \\
& Mean rank & 1.842 & 1.834 & \(-0.008\;[-0.062,0.057]\) & 1.00 \\
\addlinespace[1.5pt]
Fixed prior weights & Top-1 (\%) & 73.53 & 73.72 & \(-0.19\;[-1.56,1.33]\) & 0.82 \\
& Top-5 (\%) & 95.31 & 94.92 & \(0.39\;[-0.22,1.07]\) & 0.96 \\
& Mean rank & 1.842 & 1.906 & \(0.064\;[-0.020,0.170]\) & 1.00 \\
\end{longtable}
\endgroup

Uniform weighting gives the highest Top-1 point estimate, 74.22\% versus 73.53\% for query-aligned weighting, and all four paired Top-1 intervals span zero. Selecting the maximum-weight route yields 59.91\% Top-1, whereas the any-correct-route oracle reaches 83.62\%; mean normalized weight entropy is 0.938. Uniform summation gives a similar Top-1 estimate with the simplest weighting rule and falls within the same inferential range as the validation-fixed query-aligned configuration.

\begin{figure}[!t]
\centering
\includegraphics[width=0.94\linewidth]{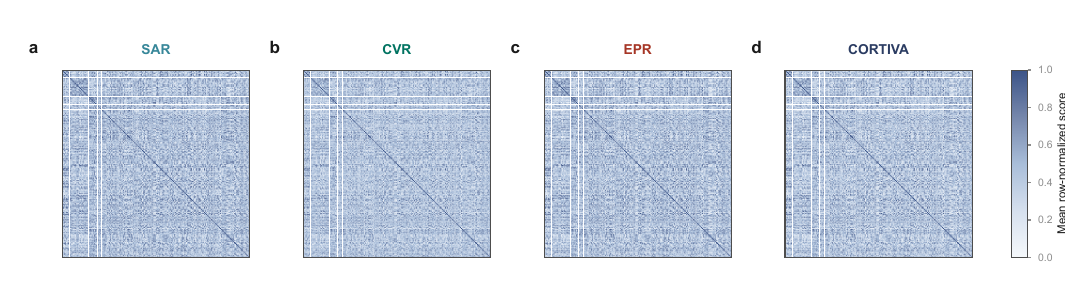}
\caption{\textbf{Route-specific candidate-score structure.}
\textbf{a--d}, Mean row-normalized score matrices for the SAR, CVR, EPR, and fused CORTIVA outputs. Queries and candidates use the same label-derived high-level group ordering; white lines mark group boundaries.}
\label{fig:supp-route-score-structure}
\end{figure}
\FloatBarrier

\Needspace{26\baselineskip}
\section{Input sensitivity and practical sampling}
\label{sec:supp-practical-sampling}

Input sensitivity is measured at the selected 60-epoch participant checkpoints, holding visual targets and fusion parameters fixed across every perturbation.

\begin{figure}[!ht]
\centering
\includegraphics[width=0.90\linewidth]{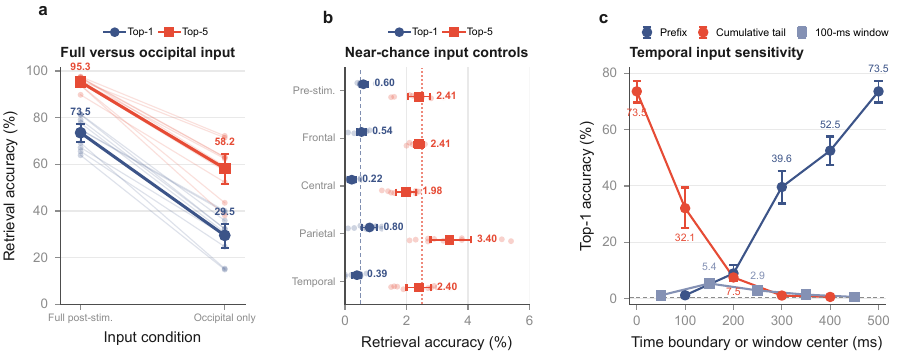}
\caption{\textbf{Sensor and temporal input-ablation sensitivity.} \textbf{a}, Paired participant-level Top-1 and Top-5 accuracy for the full 0--500-ms post-stimulus input and occipital-only input; bold trajectories connect group means. \textbf{b}, Top-1 and Top-5 accuracy for duration-matched pre-stimulus input and four additional sensor groups, with metric-specific analytical-chance levels. Small points, large markers, and whiskers in \textbf{a,b} denote participant means, group means, and 95\% intervals. \textbf{c}, Cumulative-prefix, cumulative-tail, and independent 100-ms-window Top-1 sensitivity. Participant means are computed after averaging five optimization seeds within participant. The full input uses all 63 channels; retained-input conditions preserve the named samples and set all others to zero. The pre-stimulus and frontal-only results are also summarized with the orthogonal controls in Fig.~\ref{fig:supp-e60-negative-controls}.}
\label{fig:supp-e60-frozen-input}
\end{figure}

\begin{figure}[!ht]
\centering
\includegraphics[width=0.90\linewidth]{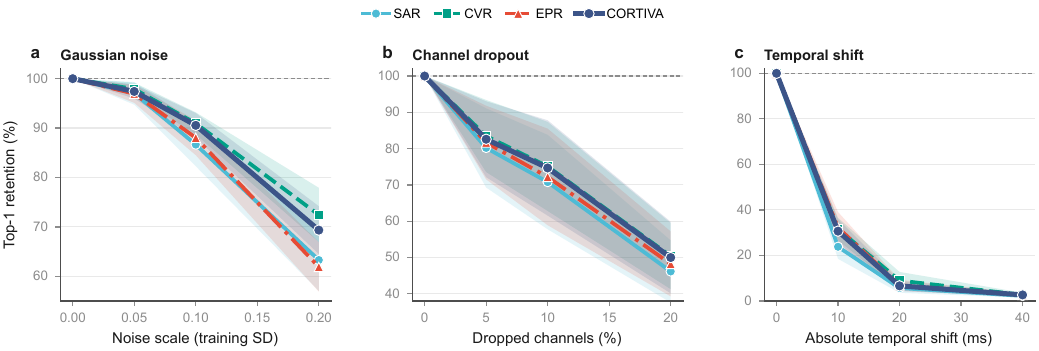}
\caption{\textbf{EEG perturbation sensitivity.} Top-1 retention relative to the clean four-observation input under \textbf{a}, Gaussian noise scaled by the participant's training-set channel standard deviations; \textbf{b}, nested channel dropout; and \textbf{c}, symmetric zero-padded temporal shifts. Four deterministic realizations are averaged for noise and channel dropout; positive and negative shifts are averaged within absolute shift.}
\label{fig:supp-e60-robustness}
\end{figure}
\FloatBarrier

\begingroup
\suppDenseTableFormat
\renewcommand{\arraystretch}{1.03}
\begin{longtable}{@{}L{0.27\textwidth}L{0.20\textwidth}C{0.12\textwidth}C{0.12\textwidth}C{0.14\textwidth}@{}}
\caption{Input-sensitivity analysis. Values average five optimization seeds within participant and then ten participants. The pre-stimulus row uses real, duration-matched input; sensor-group and temporal rows preserve the named samples at a frozen checkpoint and set all others to zero.}\label{tab:supp-frozen-input-scan}\\
\toprule
\tblhead{Condition} & \tblhead{Family} & \tblhead{Top-1 (\%)} & \tblhead{Top-5 (\%)} & \tblhead{Mean rank} \\
\midrule
\endfirsthead
\caption[]{Input-sensitivity analysis (Continued).}\\
\toprule
\tblhead{Condition} & \tblhead{Family} & \tblhead{Top-1 (\%)} & \tblhead{Top-5 (\%)} & \tblhead{Mean rank} \\
\midrule
\endhead
\bottomrule
\endlastfoot
Full 0--500 ms & Full input & 73.53 & 95.31 & 1.842 \\
Pre-stimulus input & Matched negative & 0.60 & 2.41 & 100.069 \\
Frontal sensors & Sensor group & 0.54 & 2.41 & 99.831 \\
Central sensors & Sensor group & 0.22 & 1.98 & 105.785 \\
Parietal sensors & Sensor group & 0.80 & 3.40 & 89.868 \\
Temporal sensors & Sensor group & 0.39 & 2.40 & 101.730 \\
Occipital sensors & Sensor group & 29.54 & 58.23 & 12.480 \\
0--100 ms & Cumulative prefix & 1.19 & 6.10 & 79.065 \\
0--200 ms & Cumulative prefix & 8.95 & 24.40 & 37.238 \\
0--300 ms & Cumulative prefix & 39.60 & 72.16 & 6.764 \\
0--400 ms & Cumulative prefix & 52.52 & 83.38 & 3.795 \\
100--500 ms & Cumulative tail & 32.09 & 62.67 & 10.685 \\
200--500 ms & Cumulative tail & 7.53 & 21.42 & 42.999 \\
300--500 ms & Cumulative tail & 1.10 & 4.73 & 81.331 \\
400--500 ms & Cumulative tail & 0.62 & 3.08 & 95.486 \\
0--100 ms & Independent window & 1.19 & 6.10 & 79.065 \\
100--200 ms & Independent window & 5.40 & 15.06 & 51.503 \\
200--300 ms & Independent window & 2.91 & 9.20 & 67.549 \\
300--400 ms & Independent window & 1.42 & 4.26 & 84.278 \\
400--500 ms & Independent window & 0.62 & 3.08 & 95.486 \\
\end{longtable}
\endgroup

\Needspace{28\baselineskip}
\begingroup
\suppDenseTableFormat
\begin{longtable}{@{}L{0.24\textwidth}L{0.18\textwidth}C{0.12\textwidth}C{0.12\textwidth}C{0.14\textwidth}@{}}
\caption{CORTIVA repetition and perturbation summary. Values are participant-level means after averaging seeds within participant.}\label{tab:supp-e60-practical-sensitivity}\\
\toprule
\tblhead{Analysis} & \tblhead{Level} & \tblhead{Top-1 (\%)} & \tblhead{Top-5 (\%)} & \tblhead{Mean rank} \\
\midrule
\endfirsthead
\caption[]{CORTIVA repetition and perturbation summary (Continued).}\\
\toprule
\tblhead{Analysis} & \tblhead{Level} & \tblhead{Top-1 (\%)} & \tblhead{Top-5 (\%)} & \tblhead{Mean rank} \\
\midrule
\endhead
\bottomrule
\endlastfoot
Test repetitions & 1 & 62.71 & 90.06 & 2.739 \\
& 2 & 69.27 & 93.81 & 2.108 \\
& 3 & 71.75 & 94.88 & 1.934 \\
& 4 & 73.53 & 95.31 & 1.842 \\
\addlinespace[1.5pt]
Gaussian noise & 0.05 training SD & 71.56 & 94.66 & 1.970 \\
& 0.10 training SD & 66.45 & 92.22 & 2.357 \\
& 0.20 training SD & 50.79 & 82.58 & 4.243 \\
\addlinespace[1.5pt]
Channel dropout & 5\% & 60.91 & 85.01 & 5.868 \\
& 10\% & 55.24 & 79.61 & 8.323 \\
& 20\% & 36.79 & 62.54 & 15.678 \\
\addlinespace[1.5pt]
Temporal shift & 10 ms & 22.45 & 50.31 & 15.018 \\
& 20 ms & 4.84 & 16.13 & 51.050 \\
& 40 ms & 1.96 & 6.63 & 76.122 \\
\end{longtable}
\endgroup

\begin{figure}[!ht]
\centering
\includegraphics[width=0.90\linewidth]{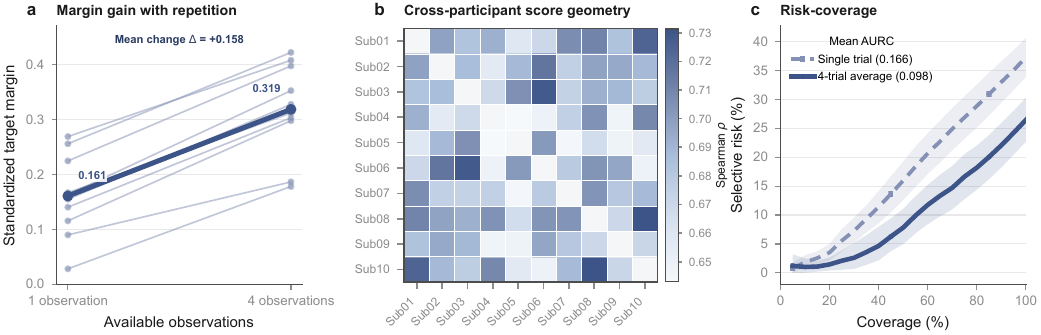}
\caption{\textbf{Repetition-dependent retrieval geometry.} \textbf{a}, Participant-paired standardized target-versus-hardest-negative margins for one and four available observations; the navy line is the participant mean, endpoint labels give its values, and the central annotation gives the mean paired change. \textbf{b}, Spearman correlations between participant-level four-observation fused-score matrices after averaging five optimization seeds. The off-diagonal mean is \(\rho=0.684\) (range 0.643--0.731), quantifying cross-participant consistency of retrieval geometry; the unit diagonal is omitted and shown as blank cells. \textbf{c}, Selective risk as retrievals are retained from highest to lowest confidence; bands are 95\% participant-bootstrap intervals, line styles distinguish one versus four observations, and legend values report participant-mean AURC.}
\label{fig:supp-e60-repetition-geometry}
\end{figure}

\Needspace{8\baselineskip}
\section{Representational analyses}

\begin{figure}[!t]
\centering
\includegraphics[width=0.94\linewidth]{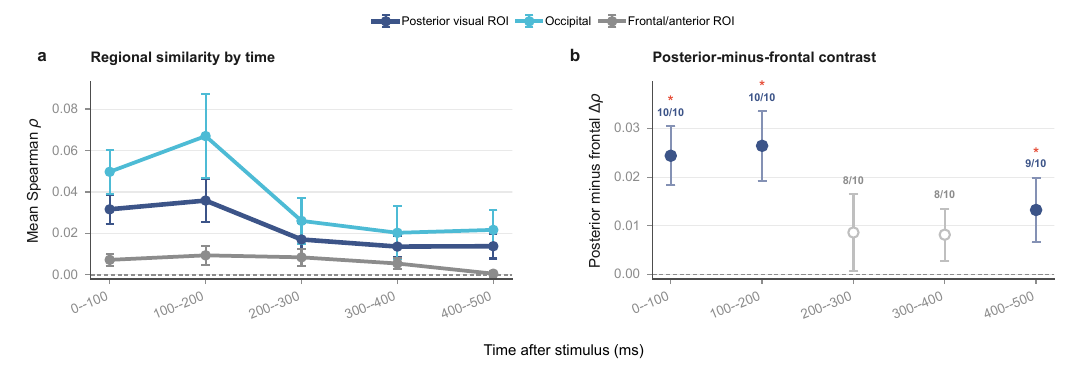}
\caption{\textbf{Regional temporal summary of EEG--CLIP representational similarity.}
\textbf{a}, Mean regional Spearman correlations across Sub01--Sub10 in five non-overlapping 100-ms windows.
\textbf{b}, Posterior-minus-frontal contrasts with 95\% \(t\)-intervals and positive-participant counts. The 0--100, 100--200, and 400--500 ms contrasts survive Holm correction over the five-window family.}
\label{fig:supp-neural-temporal-summary}
\end{figure}

\begin{figure}[!t]
\centering
\includegraphics[width=0.94\linewidth]{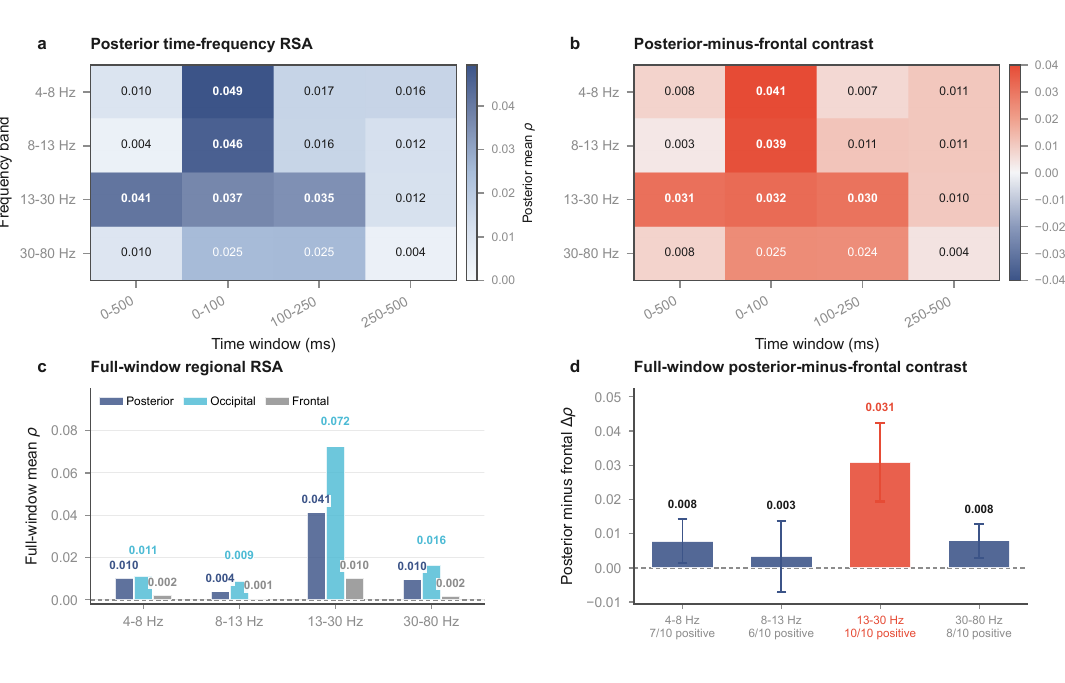}
\caption{\textbf{Time--frequency EEG--CLIP RSA.} \textbf{a}, Posterior mean correlations for 4--8, 8--13, 13--30, and 30--80~Hz log-bandpower over the tested windows. \textbf{b}, Corresponding posterior-minus-frontal mean contrasts. \textbf{c}, Full-window posterior, occipital, and frontal means. \textbf{d}, Full-window posterior-minus-frontal contrasts with 95\% \(t\)-intervals across Sub01--Sub10; labels report positive-participant counts. Twenty of 56 tested contrasts survive Benjamini--Hochberg false-discovery-rate correction. The 500-ms window provides 2-Hz native resolution; zero padding aligns the frequency grid across windows.}
\label{fig:supp-time-frequency-rsa-controls}
\end{figure}

RSA compares neural and model representational geometries \citep{Kriegeskorte2008RSA}. The mean EEG--CLIP Spearman \(\rho\) over posterior sensors is 0.047 for 0--500~ms and positive in all ten participants. O1/Oz/O2 have a full-window regional mean of \(\rho=0.064\); the posterior-minus-frontal difference is \(\Delta\rho=0.027\) (one-sided exact sign test, \(p=0.00098\)). Among the five 100-ms windows, the largest contrast occurs at 100--200 ms (\(\Delta\rho=0.026\)); the 0--100, 100--200, and 400--500 ms contrasts remain positive after Holm correction (Supplementary Figs.~\ref{fig:supp-neural-temporal-summary} and~\ref{fig:supp-time-frequency-rsa-controls}).

Time--frequency RSA uses log-bandpower in the 4--8, 8--13, 13--30, and 30--80~Hz bands on the 0--500~ms, 500-Hz analysis grid. FFTs are zero-padded to 512 points for a common frequency grid across windows; the 2-Hz native resolution of a 500-ms window makes the lowest-frequency estimates coarse.

\clearpage
\section{Training duration and optimization stability}

Validation-loss minima within the 60-epoch cap determine the participant checkpoints used for the final five-seed evaluation.

\begingroup
\suppDenseTableFormat
\begin{longtable}{@{}L{0.15\textwidth}C{0.16\textwidth}C{0.16\textwidth}C{0.12\textwidth}C{0.20\textwidth}@{}}
\caption{Validation-loss improvement with extended training. Relative improvement is measured from epoch 18 to the selected concept-disjoint validation minimum.}\label{tab:supp-epoch-budget-gate}\\
\toprule
\tblhead{Participant} & \tblhead{Epoch-18 loss} & \tblhead{Selected loss} & \tblhead{Selected epoch} & \tblhead{Relative improvement} \\
\midrule
\endfirsthead
\caption[]{Validation-loss improvement with extended training (Continued).}\\
\toprule
\tblhead{Participant} & \tblhead{Epoch-18 loss} & \tblhead{Selected loss} & \tblhead{Selected epoch} & \tblhead{Relative improvement} \\
\midrule
\endhead
\bottomrule
\endlastfoot
Sub01 & 1.0471 & 0.9671 & 60 & 7.64\% \\
Sub02 & 1.2298 & 1.1601 & 52 & 5.67\% \\
Sub03 & 1.1821 & 1.1110 & 59 & 6.02\% \\
Sub04 & 1.1725 & 1.0987 & 60 & 6.29\% \\
Sub05 & 1.3888 & 1.3413 & 53 & 3.42\% \\
Sub06 & 1.0223 & 0.9366 & 60 & 8.39\% \\
Sub07 & 1.0098 & 0.9236 & 60 & 8.53\% \\
Sub08 & 0.8270 & 0.7446 & 60 & 9.97\% \\
Sub09 & 1.1427 & 1.0772 & 58 & 5.73\% \\
Sub10 & 0.8396 & 0.7582 & 60 & 9.69\% \\
\midrule
Mean & 1.0862 & 1.0118 & -- & 7.13\% \\
\end{longtable}
\endgroup

Relative to epoch 18, the selected checkpoints reduce participant-wise validation loss by 7.13\% on average, supporting the 60-epoch schedule used for the five-seed evaluation.

\begingroup
\suppDenseTableFormat
\begin{longtable}{@{}L{0.34\textwidth}C{0.18\textwidth}C{0.18\textwidth}C{0.18\textwidth}@{}}
\caption{Multi-seed stability and participant-level uncertainty for CORTIVA on THINGS-EEG2. Seed summaries average Sub01--Sub10; participant intervals follow five-seed averaging.}\label{tab:supp-e60-budget-robustness}\\
\toprule
\tblhead{Statistic} & \tblhead{Top-1 (\%)} & \tblhead{Top-5 (\%)} & \tblhead{Mean rank} \\
\midrule
\endfirsthead
\caption[]{Multi-Seed Stability and Participant-Level Uncertainty (Continued).}\\
\toprule
\tblhead{Statistic} & \tblhead{Top-1 (\%)} & \tblhead{Top-5 (\%)} & \tblhead{Mean rank} \\
\midrule
\endhead
\bottomrule
\endlastfoot
Across-seed mean $\pm$ SD & $73.53\pm0.43$ & $95.31\pm0.25$ & $1.842\pm0.005$ \\
Across-seed range & 73.10--74.00 & 95.05--95.70 & 1.837--1.849 \\
Participant mean [95\% CI] & 73.53 [69.62, 77.33] & 95.31 [93.81, 96.56] & 1.842 [1.654, 2.083] \\
\end{longtable}
\endgroup

\FloatBarrier
\Needspace{25\baselineskip}
\section{THINGS-MEG temporal and sensor analyses}

\begin{figure}[!ht]
\centering
\includegraphics[width=0.94\linewidth]{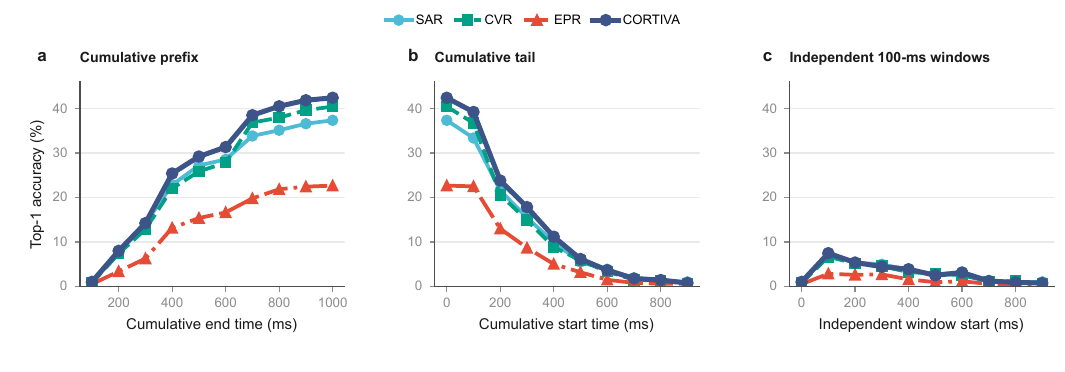}
\caption{\textbf{THINGS-MEG temporal sensitivity.} \textbf{a}, Cumulative-prefix Top-1 accuracy as the input expands from 0 to 1000 ms. \textbf{b}, Cumulative-tail accuracy as progressively earlier samples are removed. \textbf{c}, Independent 100-ms windows. Curves show four-participant means after averaging six optimization seeds.}
\label{fig:supp-meg-temporal}
\end{figure}

\begin{figure}[!ht]
\centering
\includegraphics[width=\linewidth]{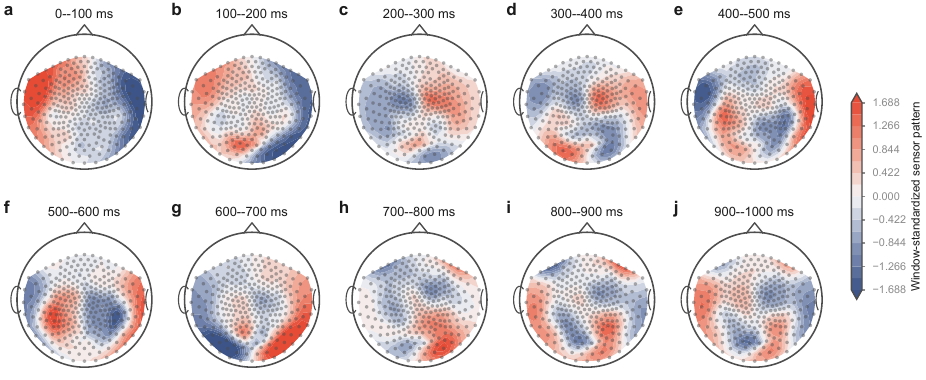}
\caption{\textbf{THINGS-MEG sensor-pattern topography.} The ten panels show the mean within-participant, window-standardized sensor pattern across four participants and 271 magnetometers in non-overlapping 100-ms windows from 0--100 to 900--1000~ms. All panels share one zero-centered scale and use the recorded MEG sensor geometry.}
\label{supp-fig:supp-meg-evoked-topography}
\end{figure}

\FloatBarrier
\Needspace{18\baselineskip}
\section{Independent DINOv2 validation}

DINOv2 ViT-B/14 \citep{Oquab2024DINOv2} is used only for external visual-geometry analyses. Its fixed features for the 200 held-out images support error-neighborhood analysis and EEG representational-similarity analysis. Error-neighborhood rates pool incorrect outcomes over five seeds within each participant before participant-level inference. Fixed 0--500-ms EEG tensors and repeated measurements keep this external validation separate from model fitting and score selection.

\begingroup
\suppDenseTableFormat
\renewcommand{\arraystretch}{1.08}
\begin{longtable}{@{}L{0.24\textwidth}C{0.13\textwidth}C{0.18\textwidth}L{0.22\textwidth}C{0.15\textwidth}@{}}
\caption{Independent DINOv2 error-geometry and EEG representational-similarity validation. Intervals are 95\% participant-level intervals.}
\label{tab:supp-dinov2-validation}\\
\toprule
\tblhead{Analysis} & \tblhead{Estimate} & \tblhead{95\% CI} & \tblhead{Reference or contrast} & \tblhead{Participants; \(p\)} \\
\midrule
\endfirsthead
\caption[]{Independent DINOv2 error-geometry and EEG representational-similarity validation (Continued).}\\
\toprule
\tblhead{Analysis} & \tblhead{Estimate} & \tblhead{95\% CI} & \tblhead{Reference or contrast} & \tblhead{Participants; \(p\)} \\
\midrule
\endhead
\bottomrule
\endlastfoot
\multicolumn{5}{l}{\textbf{A. DINOv2 neighborhood of incorrect Top-1 predictions}} \\
Top-10 error-neighbor rate & 21.93\% & 18.80--25.06\% & Random wrong: 5.03\%; 4.36\(\times\) & 10/10; 0.00098 \\
Top-25 error-neighbor rate & 33.87\% & 29.44--38.30\% & Random wrong: 12.56\%; 2.70\(\times\) & 10/10; 0.00098 \\
Top-50 error-neighbor rate & 45.77\% & 41.17--50.38\% & Random wrong: 25.13\%; 1.82\(\times\) & 10/10; 0.00098 \\
\midrule
\multicolumn{5}{l}{\textbf{B. EEG--DINOv2 representational similarity}} \\
Posterior, 0--500 ms & \(\rho=0.0316\) & 0.0258--0.0375 & Spearman RSA against zero & 10/10; 0.00098 \\
Posterior minus frontal & \(\Delta\rho=0.0108\) & 0.0067--0.0149 & Full-window regional contrast & 10/10; 0.00098 \\
Posterior late minus early & \(\Delta\rho=0.0159\) & 0.0072--0.0247 & 250--500 vs. 0--100 ms & 9/10; 0.0107 \\
Posterior split-half reliability & \(\rho=0.1505\) & 0.1066--0.1944 & Repeated-measure split halves & -- \\
\end{longtable}
\endgroup

\FloatBarrier
\Needspace{26\baselineskip}
\section{Additional visual analyses}

\begin{figure}[!ht]
\centering
\includegraphics[width=0.92\linewidth]{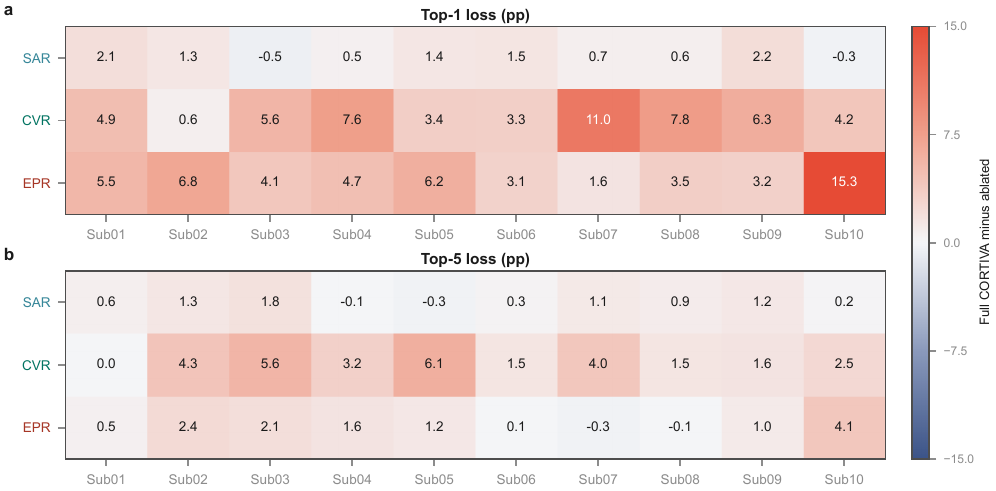}
\caption{\textbf{Participant-level consistency of matched route-removal effects.} \textbf{a}, Top-1 loss after removing each route. \textbf{b}, Top-5 loss under the same matched retraining protocol. Each cell is the full CORTIVA participant mean minus the corresponding ablated participant mean after averaging five optimization seeds; positive values indicate a retrieval loss after route removal.}
\label{supp-fig:supp-e60-ablation-profile}
\end{figure}

\begin{figure}[!ht]
\centering
\includegraphics[width=0.94\linewidth]{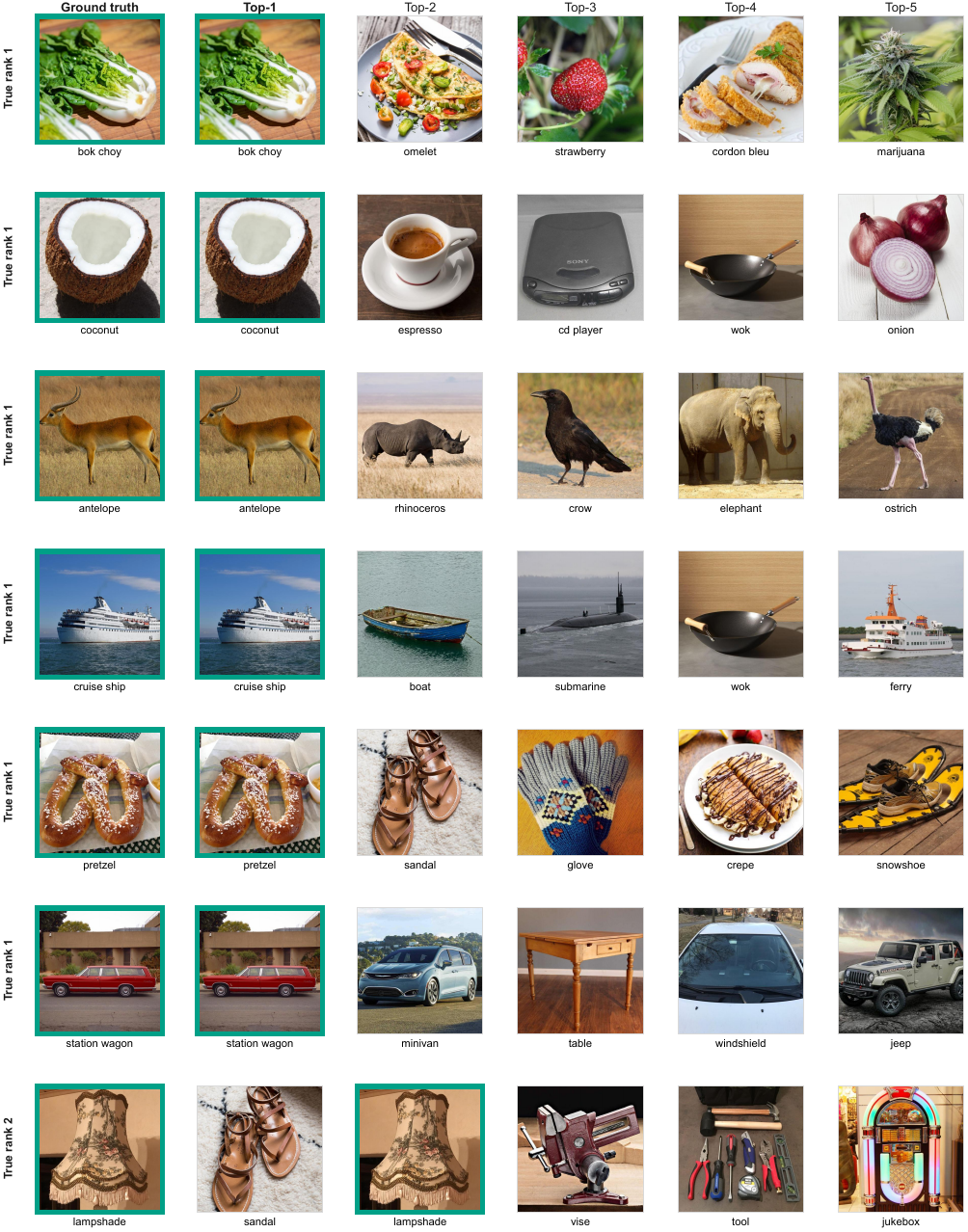}
\caption{\textbf{Representative retrievals.} Six Top-1 outcomes and one Top-2 outcome from Sub10 under seed 2026, selected to span score quantiles and semantic categories. Each row shows the ground-truth image followed by the five highest-scoring candidates; the teal border marks the correct candidate. Stimulus source: the THINGS object-concept image database \citep{Hebart2019THINGS}.}
\label{fig:supp-e60-retrieval-gallery}
\end{figure}
\FloatBarrier

\begin{figure}[!t]
\centering
\includegraphics[width=\linewidth]{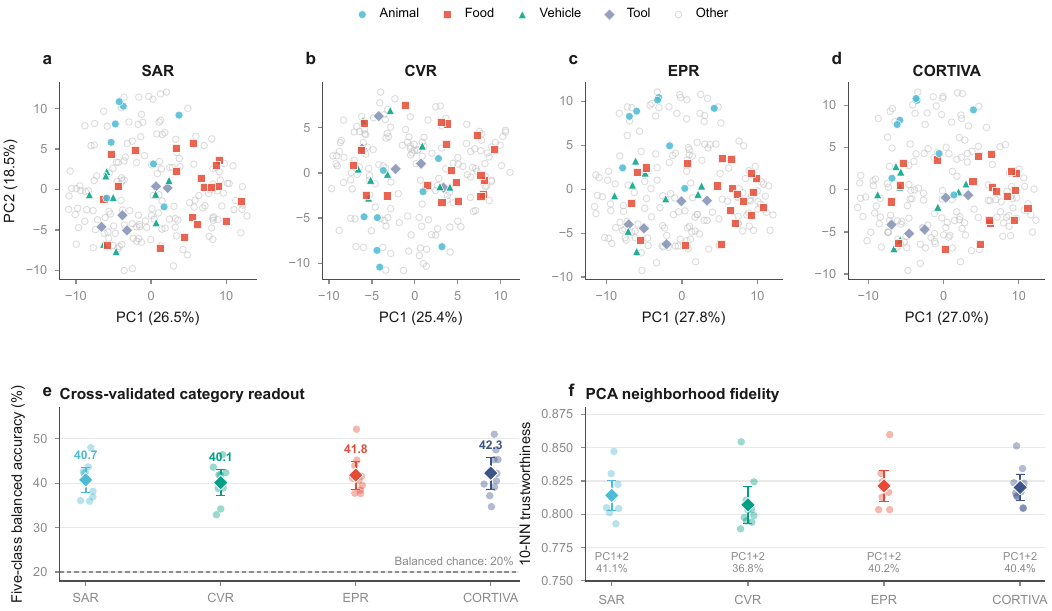}
\caption{\textbf{Score-space projection and broad-category readout.}
\textbf{a--d}, Principal-component projections of SAR, CVR, EPR, and CORTIVA score vectors after row standardization within participant and averaging over Sub01--Sub10. Each point is one held-out query; all 200 concepts are shown, and category labels are not used to fit PCA.
\textbf{e}, Five-class balanced accuracy from five repeats of stratified five-fold cross-validation within each participant. The training-fold pipeline standardizes features, retains 95\% variance by PCA, and fits class-balanced logistic regression.
\textbf{f}, Ten-neighbor trustworthiness of each participant-specific two-dimensional PCA; labels give the participant-mean variance explained by PC1 and PC2. In \textbf{e,f}, pale points denote participants and diamonds with bars show participant means and 95\% \(t\)-intervals.}
\label{fig:supp-score-space-projection}
\end{figure}

Repeated five-fold category readout reaches 40.1--42.3\% balanced accuracy across routes against a 20\% balanced chance level. The corresponding cosine silhouette coefficients (-0.17 to -0.15) characterize a diffuse, distributed geometry alongside strong linear category readout. Supplementary Fig.~\ref{fig:supp-score-space-projection} visualizes this coexistence across all 200 test queries.

\clearpage
\makeatletter
\setlength{\@fptop}{0pt}
\setlength{\@fpbot}{0pt plus 1fil}
\makeatother
\FloatBarrier

\subsection*{Acknowledgment}

The authors thank Peng Xie, Xilin Tao, Bowen Gong, and Xingze Chen for insightful discussions and constructive suggestions that sharpened the presentation of this work.

\end{document}